\documentclass[10pt,twocolumn,letterpaper]{article}

\usepackage{color}
\definecolor{Silver}{rgb}{0.8,0.8,0.8}     
\usepackage[pagenumbers]{cvpr} 

\usepackage[dvipsnames]{xcolor}
\usepackage{amsmath}
\usepackage{verbatim}
\usepackage{tabularray}
\usepackage{rotating}
\usepackage{array}
\usepackage{rotating}
\usepackage{multirow}
\usepackage{ragged2e}
\usepackage{scalerel}
\usepackage{bbm}
\usepackage{graphicx}
\usepackage{wrapfig}
\usepackage{booktabs}
\usepackage{xspace}

\newcommand{\subheader}[1]{\vspace{0.25\baselineskip} \noindent \textbf{#1.}}

\newcommand{\smaller}{\fontsize{8pt}{9pt}\selectfont}

\newcommand{\pullup}{\vspace{-0.2\baselineskip}}
\newcommand{\pullupp}{\vspace{-0.4\baselineskip}}
\newcommand{\pulluppp}{\vspace{-0.8\baselineskip}}

\newcommand{\skipless}{
	\setlength{\abovedisplayskip}{3.75pt}
	\setlength{\belowdisplayskip}{3.75pt}
	\setlength{\abovedisplayshortskip}{3.75pt}
	\setlength{\belowdisplayshortskip}{3.75pt}
}
\newcommand{\skipnormal}{
	\setlength{\abovedisplayskip}{8pt}
	\setlength{\belowdisplayskip}{8pt}
	\setlength{\abovedisplayshortskip}{5pt}
	\setlength{\belowdisplayshortskip}{8pt}
}

\definecolor{cvprblue}{rgb}{0.21,0.49,0.74}
\usepackage[pagebackref,breaklinks,colorlinks,citecolor=cvprblue]{hyperref}

\def\paperID{0016} 
\def\confName{CVPR}
\def\confYear{2024}

\title{Beyond Deepfake Images: Detecting AI-Generated Videos}

\author{Danial Samadi Vahdati, Tai D. Nguyen, Aref Azizpour, Matthew C. Stamm\\
Drexel University\\
Philadelphia, PA, USA\\
{\tt\small \{ds3729,tdn47,aa4639,mcs382\}@drexel.edu}
}

\begin{document}
\maketitle

\begin{abstract}

Recent advances in generative AI have led to the development of techniques to generate visually realistic synthetic video. While a number of techniques have been developed to detect AI-generated synthetic images, in this paper we show that synthetic image detectors are unable to detect synthetic videos. We demonstrate that this is because synthetic video generators introduce substantially different traces than those left by image generators. Despite this, we show that synthetic video traces can be learned, and used to perform reliable synthetic video detection or generator source attribution even after H.264 re-compression. Furthermore, we demonstrate that while detecting videos from new generators through zero-shot transferability is challenging, accurate detection of videos from a new generator
can be achieved through few-shot learning.

\end{abstract}

\section{Introduction}


In recent years, substantial progress in generative AI has produced numerous techniques for generating visually realistic synthetic images. These advances have also introduced significant  misinformation and disinformation threats.
Synthetic images can be easily produced and used as falsified visual evidence to deceive a target audience. 

To combat this, researchers have developed a number of techniques to detect synthetic images.  
These techniques operate by searching for statistical traces left in synthetic images by their source generator. 
%
%
For example, prior work by Zhang et al. has shown that the upsampling operation used in many generator architectures to grow an image from a small latent representation to a full sized image leaves behind traces similar to those left by resampling~\cite{resnet34_method}.  A number of approaches have been successfully developed to accurately detect synthetic images made by a wide variety of generators~\cite{luisa,genimage,swin_method2,resnet_method_2,resnet34_method2,resnet34_method3,resnet50_method4,vgg16_method3,xception_method3} and attribute them to their source~\cite{dif,genimage,openset}.


\begin{figure}[!t]
	\centering
	
	\begin{minipage}[t]{1.0\textwidth}
		\setlength{\fboxrule}{0pt}
        \setlength{\fboxsep}{0pt}
		\fbox{\parbox{0.160\textwidth}{\centering\smaller Sora}}
		\fbox{\parbox{0.160\textwidth}{\centering\smaller Stable Video Diffusion}}
		\fbox{\parbox{0.160\textwidth}{\centering\smaller Pika}}
		\smallskip
	\end{minipage}
	
	\begin{minipage}[t]{1.0\textwidth}
		\setlength{\fboxrule}{0pt}
        \setlength{\fboxsep}{0pt}
		\fbox{\centering\includegraphics[width=2.8cm, height=2.1cm]{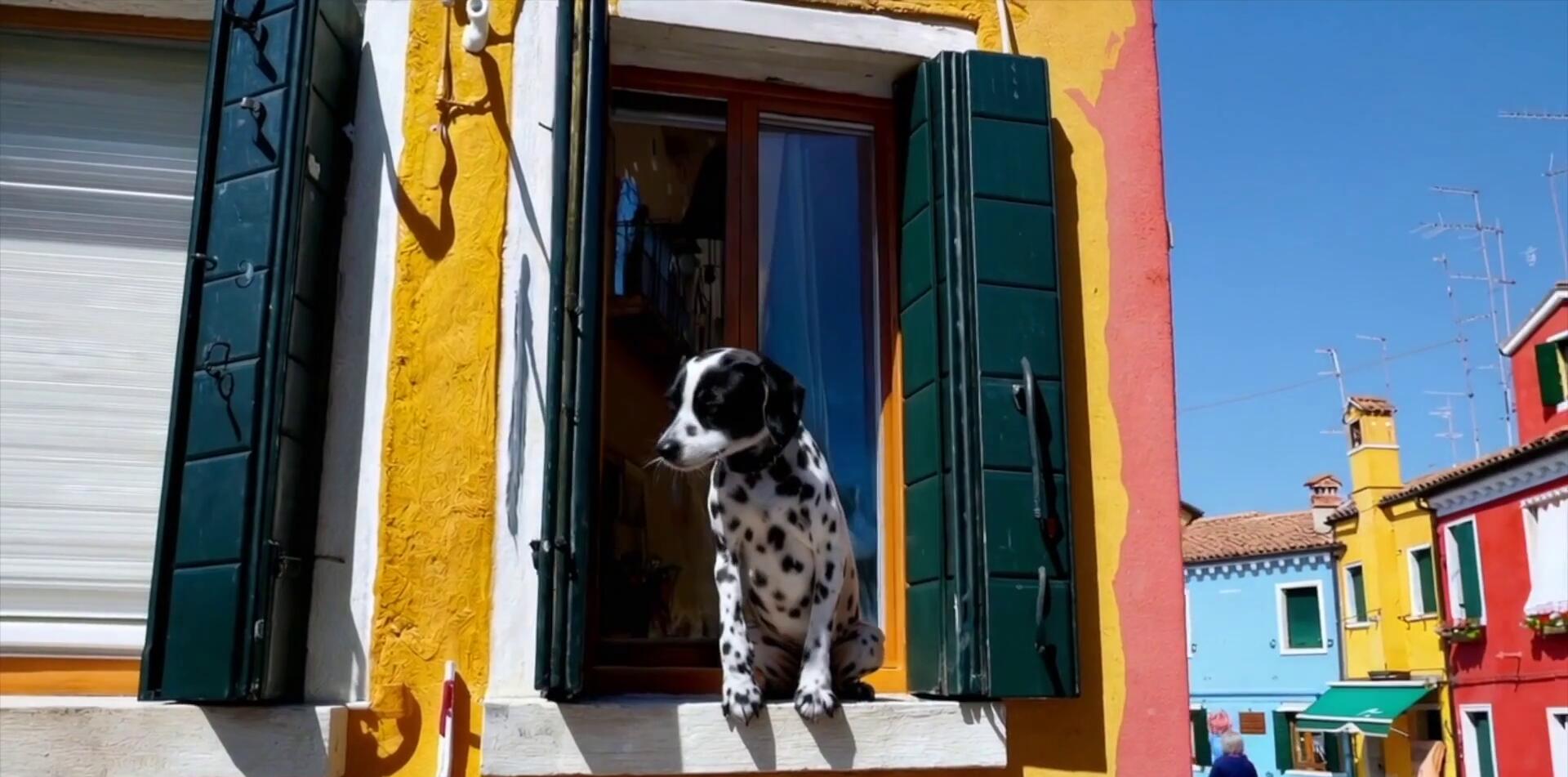}}
		\fbox{\centering\includegraphics[width=2.8cm, height=2.1cm]{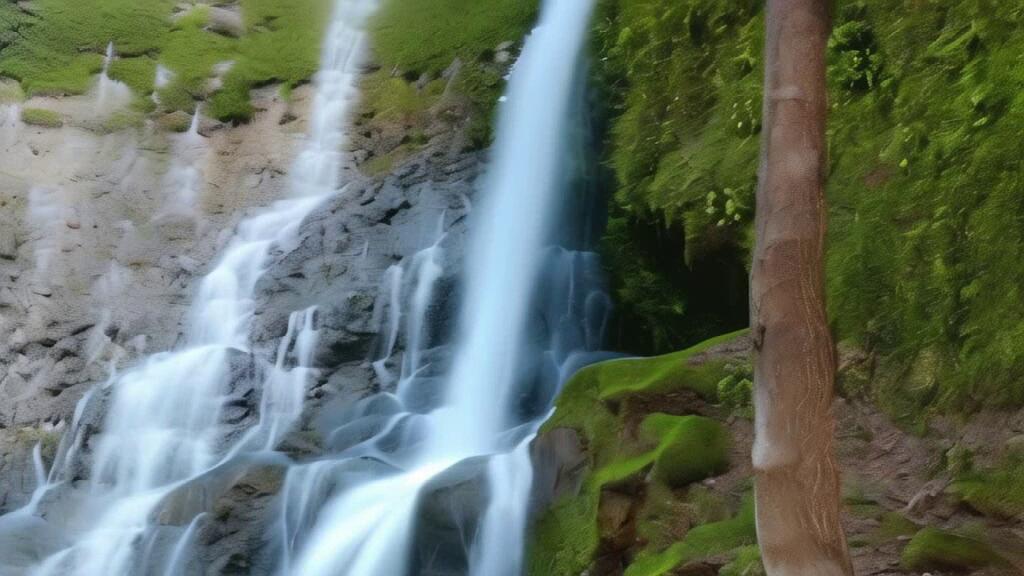}}
		\fbox{\centering\includegraphics[width=2.8cm, height=2.1cm]{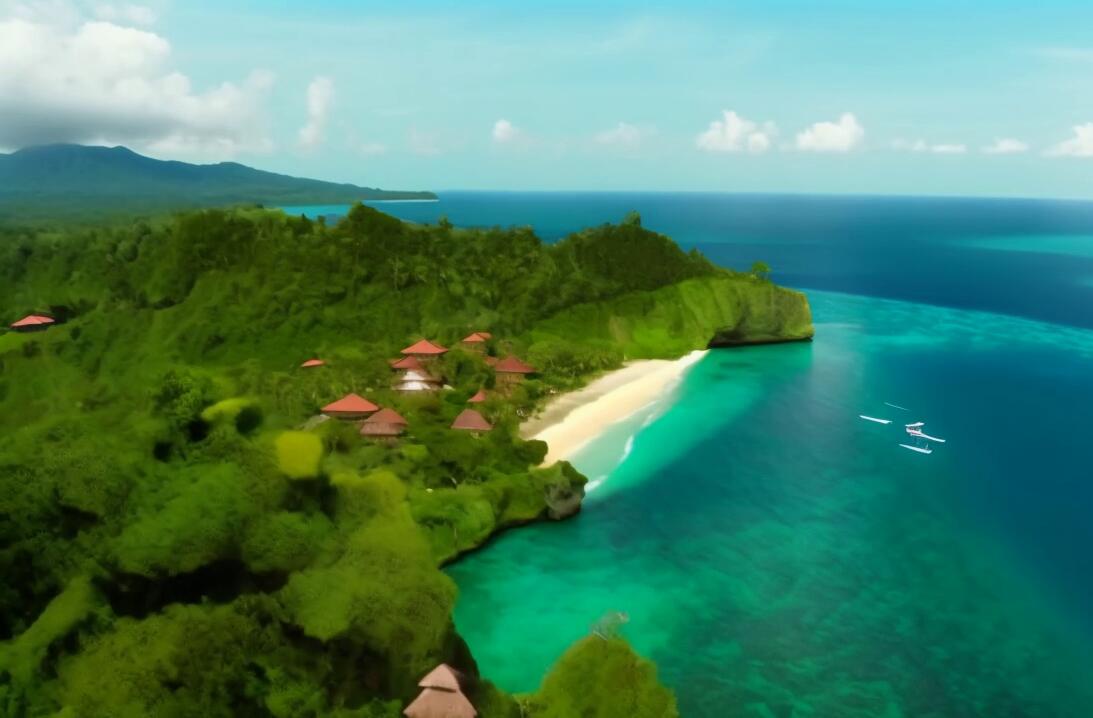}}
        \smallskip
	\end{minipage}
 
    \begin{minipage}[t]{1.0\textwidth}
		\setlength{\fboxrule}{0pt}
        \setlength{\fboxsep}{0pt}
		\fbox{\centering\includegraphics[width=2.8cm, height=2.1cm]{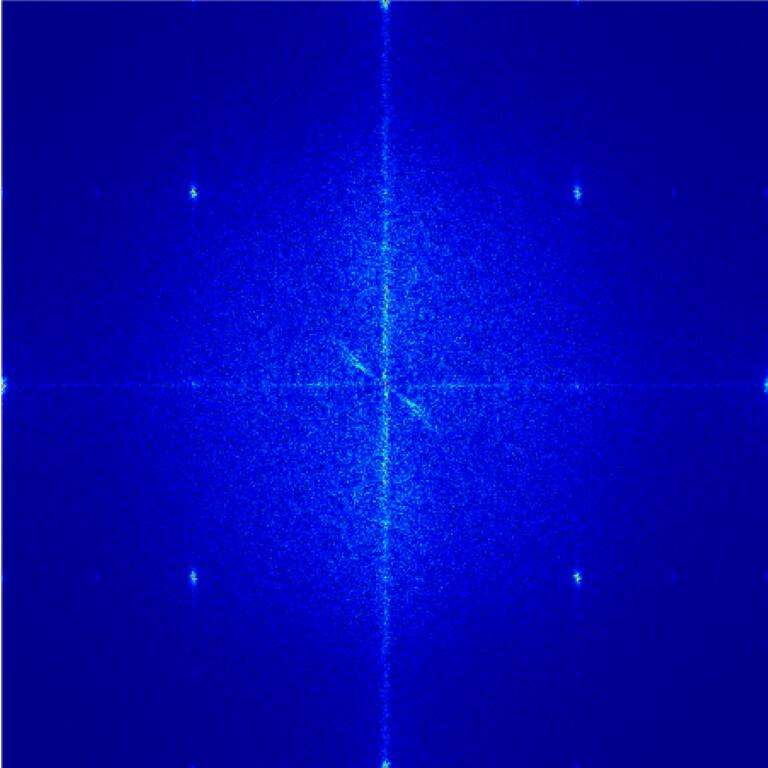}}
		\fbox{\centering\includegraphics[width=2.8cm, height=2.1cm]{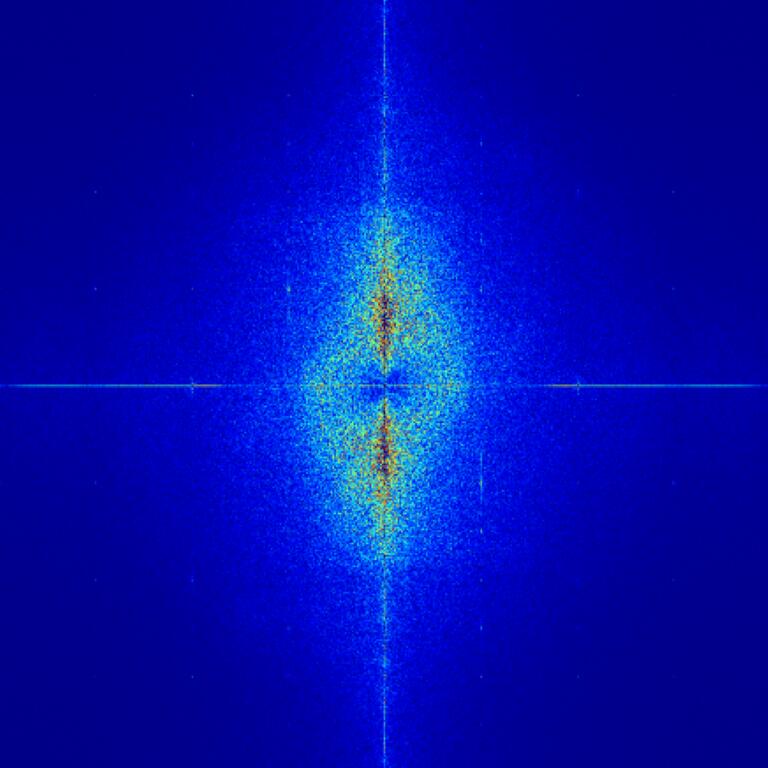}}
  		\fbox{\centering\includegraphics[width=2.8cm, height=2.1cm]{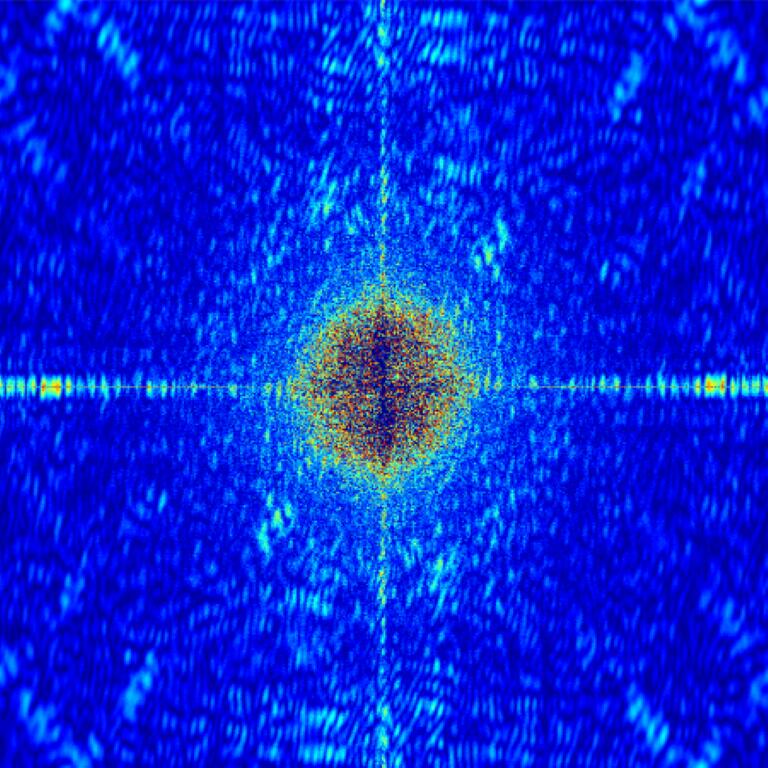}}
	\end{minipage}
 
	\caption{Top row: video frames taken from AI-generated videos. Bottom row: Fourier transforms of the residual forensic traces for each corresponding frame above. The process to produce visual results in the bottom row is described in Sec.~\ref{sec:synth_video_traces}.}
	\label{fig:front_page_graphic}
    \pulluppp\pullupp
\end{figure}

Very recently, AI-based synthetic video generators have begun to emerge.  These range from text-prompted approaches such as Stable Video Diffusion, VideoCrafter, or OpenAI's recently released Sora, 
to others such as Luma AI's NeRF-basesd approach which allows synthetic videos to be generated and manipulated based on a set of input images.  
The emergence of synthetic video generators represents not only a major technological advancement, but also a significant escalation in the potential misinformation and disinformation threats caused by generative AI.

One would reasonably assume that synthetic image detectors should accurately detect synthetic videos. 
In this paper, however, we demonstrate that synthetic image detectors do not accurately detect synthetic videos.
Furthermore, we show that this is not due to performance degradation caused by H.264 compression. Instead, we demonstrate that synthetic video generators leave distinct traces that are not detected by image detectors. 
Encouragingly, we show that these traces can be learned and utilized to perform accurate synthetic video detection and generator source attribution. 
In addition, we investigate the transferability of synthetic video detectors 
and show that they can be adapted to detect videos from new generators that contain substantially different traces using very little data.  
The novel contributions of this paper are listed below:


\begin{figure*}[!t]
    \centering
    \setlength{\fboxsep}{0pt}
    \setlength{\fboxrule}{0pt}
    
    \begin{minipage}[t]{1.0\textwidth}
        \centering
        
        
        
        \fbox{\includegraphics[height=2.4cm]{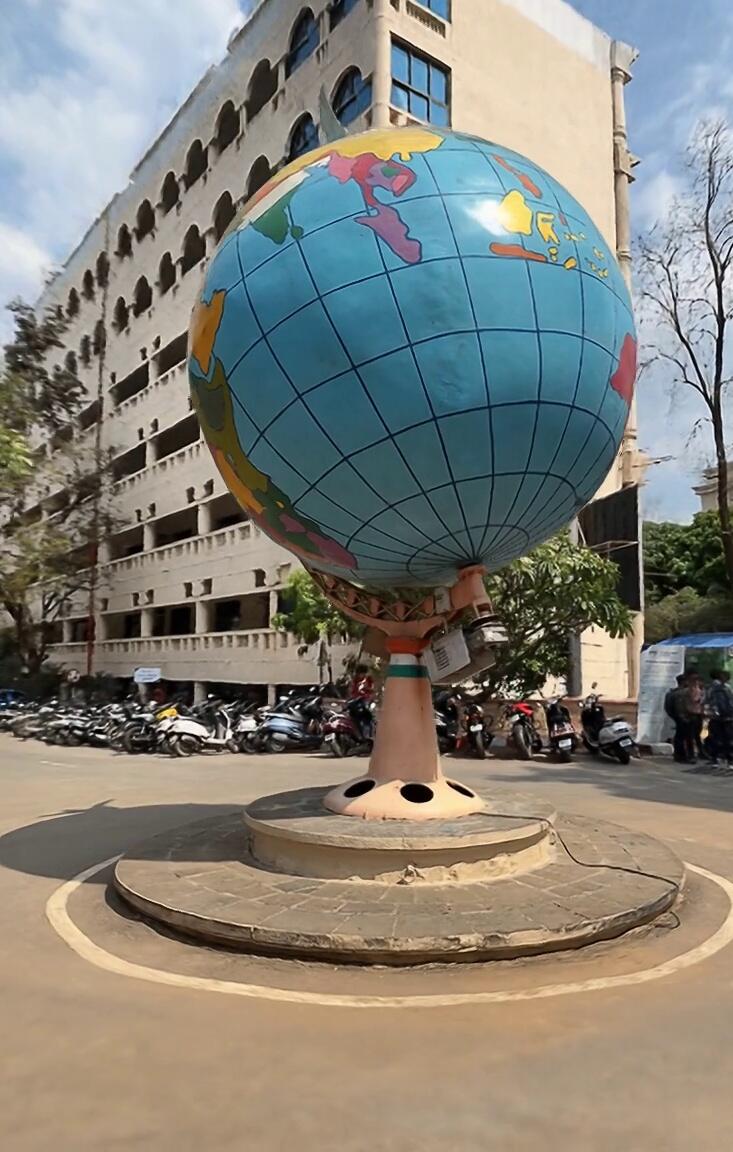}}
        \fbox{\includegraphics[height=2.4cm]{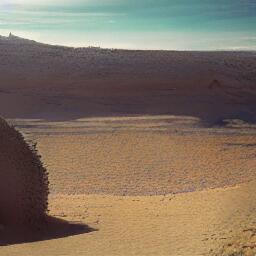}}
        \fbox{\includegraphics[height=2.4cm]{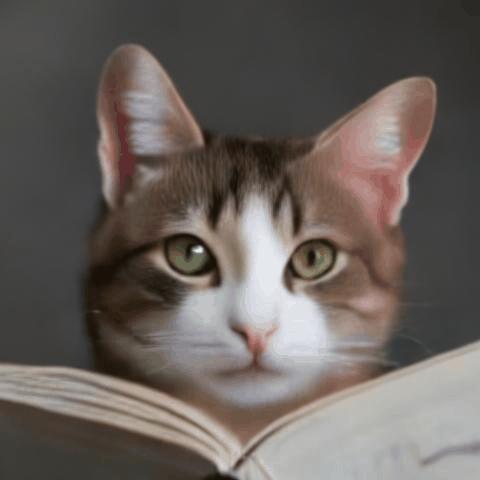}}
        \fbox{\includegraphics[height=2.4cm,trim={4cm 0 2cm 0},clip]{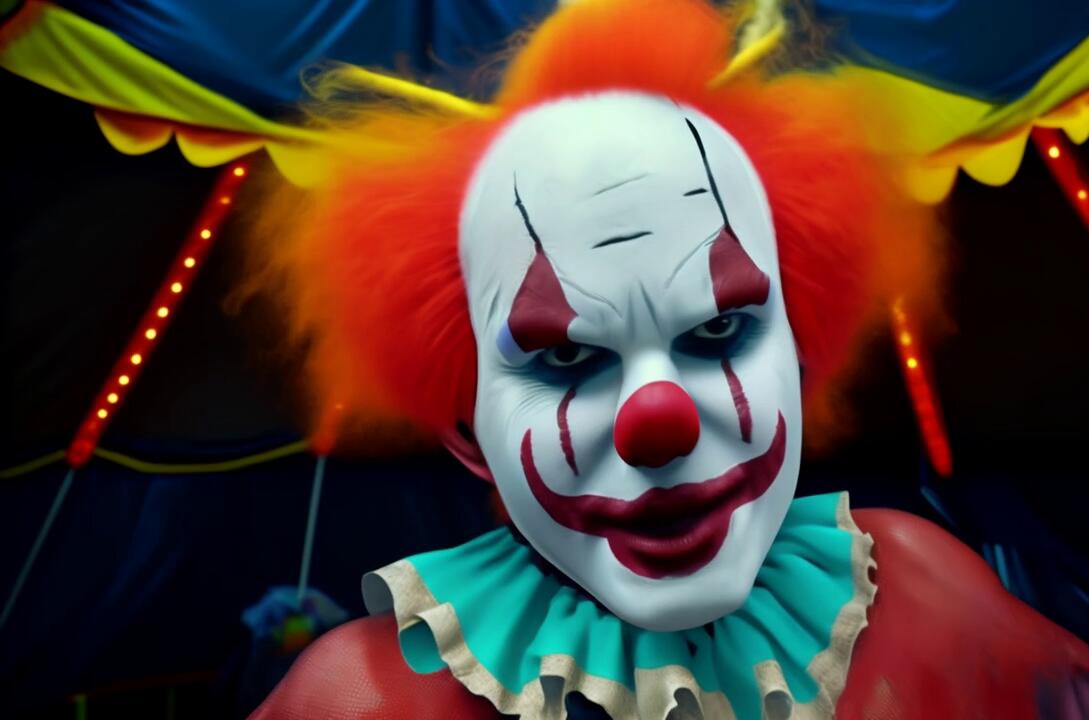}}
        \fbox{\includegraphics[height=2.4cm,trim={4.5cm 0 8.5cm 0},clip]{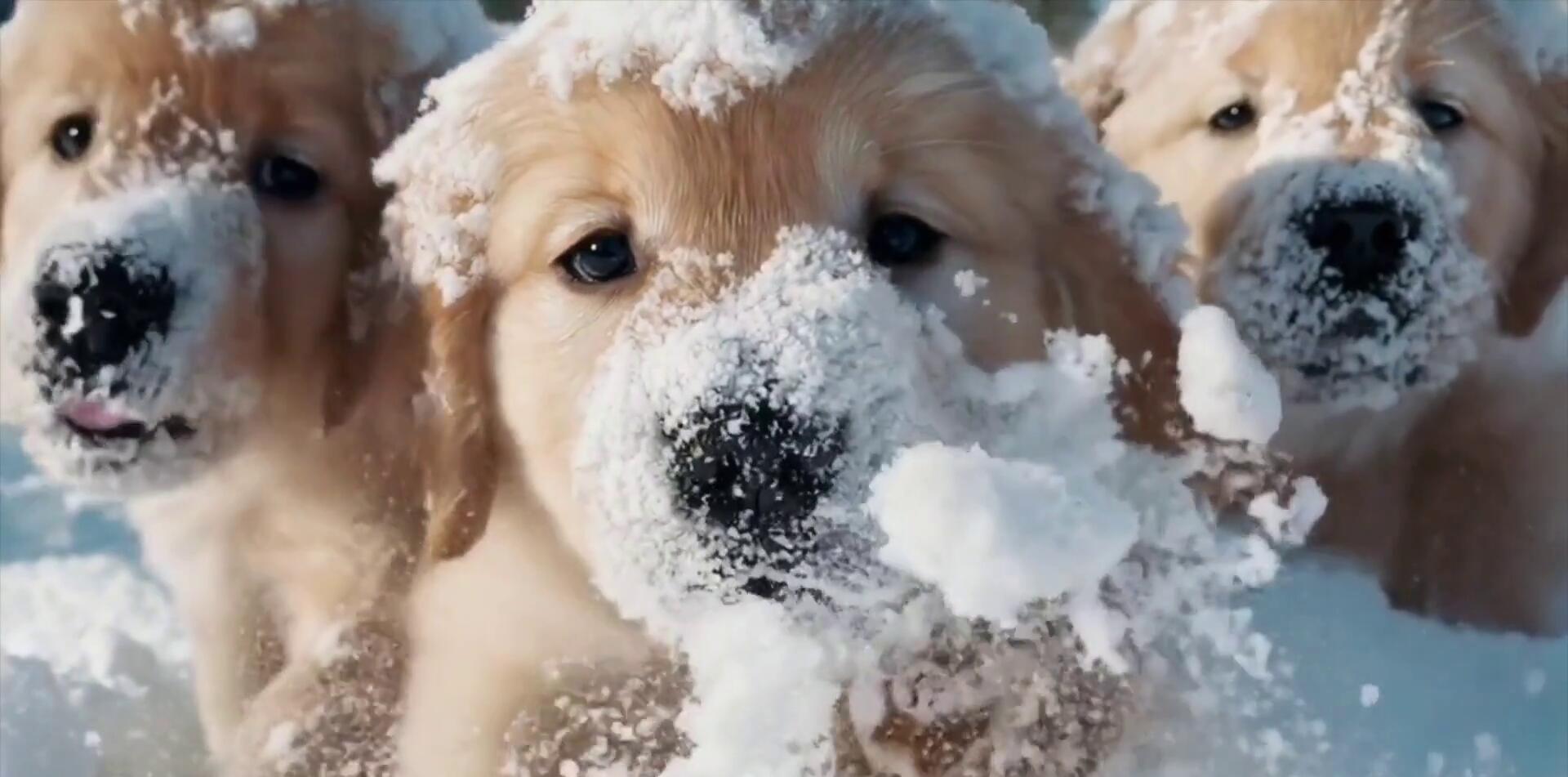}}
        \fbox{\includegraphics[height=2.4cm,trim={5cm 0 0 0},clip]{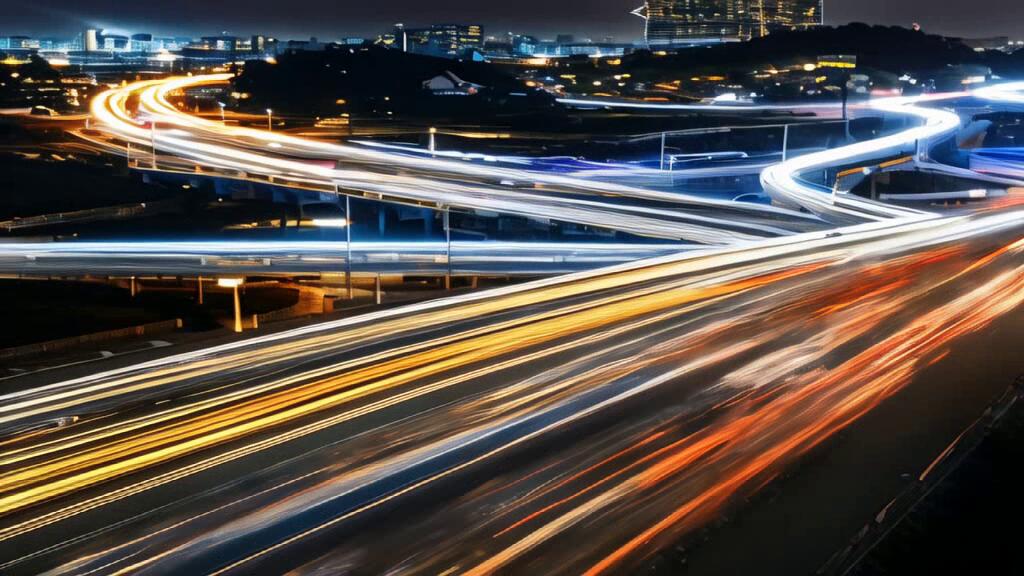}}
    \end{minipage}
    
    \caption{Sample frames from different video generators. The figure shows the synthetic video frames: from left to right Luma~\cite{Luma}, VideoCrafter~\cite{videocrafter1}, CogVideo~\cite{cogvideo}, Pika~\cite{Pika}, Sora~\cite{sora}, and Stable Video Diffusion~\cite{stablevideo}.}
    \label{fig:video_gen_visualization}
    \pulluppp\pullup
\end{figure*}

\begin{enumerate}
    \item We show that synthetic image detectors do not reliably detect AI-generated videos, and  empirically verify this is not due to the degradation effects of H.264 compression.

    \item We demonstrate that synthetic video generators leave substantially different forensic traces than those left by synthetic image generators. This is the primary cause of synthetic image detectors' poor performance on video.


    \item Furthermore, we show that synthetic video traces can be learned and used to perform reliable synthetic video detection or 
    source attribution
    even in the presence of H.264 re-compression.
    


    \item We demonstrate that while detecting videos from new generators through zero-shot transferability is challenging, accurate detection of videos from a new generator can be achieved through few-shot learning.

    \item We create a new,  
    publicly available 
    dataset of synthetic videos from a number of state-of-the-art video generators that can be used to train and benchmark the performance of synthetic video detectors.~\footnote{Link to our dataset: https://huggingface.co/datasets/ductai199x/synth-vid-detect}
    
\end{enumerate}


\section{Background}
\label{sec:background}
\subheader{Synthetic Image Generation}
The field of computer-generated media has seen significant advancements, beginning with the introduction of Generative Adversarial Networks (GANs) by Goodfellow et al.~\cite{goodfellow2014generative}, a seminal work that has since spurred a multitude of subsequent innovations~\cite{stylegan,stylegan2,stylegan3,stargan,may2023comprehensive,progan,cyclegan,vahdati2023detecting}. 
These innovations have significantly enhanced the capabilities of generative models in producing images that are diverse, realistic, and of high quality. Recent works have explored using Transformers for improving generated image consistency~\cite{transformer1,transformer2,transformer3,transformer4,transformer5,transformer6}.
However, a notable milestone was achieved with the advent of the diffusion model by Ho et al.~\cite{ho2020denoising}, which has since fueled a vast array of research leading to cutting-edge generation methods like Stable Diffusion~\cite{stablediffusion}, DALL·E~\cite{dalle2}, Midjourney~\cite{Midjourney}, and Cascade Diffusion~\cite{ho2022cascaded}, to name a few~\cite{controlnet,haji2023elasticdiffusion, diff_beat_gans, cond_text_im_gen_diff, im_gen_3d_diff}. 

\subheader{Synthetic Video Generation}
Another modality of synthetic media synthesis is synthetic video generation. Recently, lots of research attention has been devoted to developing synthetic video generation methods. These methods ranges from diffusion models~\cite{stablevideo,videocrafter1,videocrafter2}, to Transformers~\cite{cogvideo,trans1,trans2,trans3,trans4,trans5,trans6,trans7,trans8}. Moreover, there exists generation techniques that use a combination of methods such as SORA by OpenAI~\cite{sora}, and commercially available products that do not disclose the exact method used for content generation~\cite{Pika,Runway}.

\subheader{Synthetic Image Detection}
As synthetic image generators proliferate, researchers aim to devise detection methods. Wang et al.~\cite{easydetect} were among the first to tackle this by training a CNN on a single generator, enabling the detection and classification of numerous synthetic images. Subsequently, as generators grew more complex, researchers developed sophisticated detectors, including methods proposed by Marra et al.~\cite{gan1,densenet_method}, Zhang et al.~\cite{resnet34_method}, and others~\cite{vgg16_method3,gan4,gan5,gan7,monjezi,gan8,gan9,gan10,gan11,openset}. Recently, detection methods have extended to newer image generation techniques like diffusion models~\cite{luisa,dif,genimage,diff1,diff2,diff4}.



\section{Using Synthetic Image Detectors On Video}
\label{sec:synth_img_det_on_videos}

Given that a video can be seen as a sequence of images, it is reasonable to expect that synthetic image detectors should be effective at detecting AI-generated synthetic videos. Surprisingly, however, we have found that synthetic image detectors do not successfully identify synthetic videos. Furthermore, we have found 
that this issue is not primarily caused by the degradation of forensic traces due to H.264 video compression.

To demonstrate these findings, we conducted a series of experiments in which we evaluated synthetic image detector's ability to detect synthetic videos.  
The details of these experiments, as well as their outcomes, are presented below.

\begin{table}[!t]
    \centering
    \resizebox{0.98\linewidth}{!}{%
    \begin{tblr}{
        width = \linewidth,
        colspec = {m{12.5mm}m{24mm}m{33mm}m{25mm}},
        column{1} = {c},
        cell{1}{2} = {c=3}{0.893\linewidth,c},
        vline{1} = {2}{},
        vline{2-5} = {1,2}{},
        vline{-} = {3-13}{},
        hline{1-2} = {2-4}{},
        hline{2,3,6,9,14} = {-}{},
    }
        & {\textbf{Detection Algorithms}\\(P=Pretrained, R=Retrained, T=Trained-by-us)} &  & \\
        \textbf{Training} & \textbf{Refer to as} & \textbf{Architecture} & \textbf{Used by}\\
        P & Corvi et al.\cite{luisa} & ResNet-50\cite{resnet_archit} & \cite{luisa,resnet_method_2,resnet50_method_3,resnet50_method4,xception_method3}\\
        P & Sinitsa et al.\cite{dif} & DIF\cite{dif} & \cite{dif,swin_method2,genimage}\\
        P & Zhu et al.\cite{genimage} & Swin-Transformer\cite{swint} & \cite{genimage}\\
        R & ResNet-50\cite{resnet_archit} & ResNet-50\cite{resnet_archit} & \cite{luisa,resnet_method_2,resnet50_method_3,resnet50_method4,xception_method3}\\
        R & DIF\cite{dif} & DIF\cite{dif} & \cite{dif}\\
        R & Swin-T\cite{swint} & Swin-Transformer\cite{swint} & \cite{genimage,swin_method2}\\
        T & ResNet-34\cite{resnet_archit} & ResNet-34\cite{resnet_archit} & \cite{resnet34_method,resnet_method_2,resnet34_method2,resnet34_method3,xception_method3}\\
        T & VGG-16\cite{vgg16_archit} & VGG-16\cite{vgg16_archit} & \cite{vgg16_method,vgg16_method2,vgg16_method3,vgg16_method4,xception_method3}\\
        T & Xception\cite{xception_archit} & Xception\cite{xception_archit} & \cite{xception_method,xception_method2,xception_method3,xception_method4,xception_method5}\\
        T & DenseNet\cite{densenet_archit} & DenseNet\cite{densenet_archit} & \cite{densenet_method,densenet_method2,densenet_method3,densenet_method4}\\
        T & MISLnet\cite{openset} & MISLnet\cite{bayar2016deep,bayar2018constrained} & \cite{openset,mislnet_method1}
    \end{tblr}
    }
    \caption{List of detection algorithms used in this paper and the names as they are referred to in our paper.}
    \label{tab:list_of_detectors}
    \pulluppp
\end{table}

\subsection{Experimental Setup}
\label{subsec:synth_img_exp_setup}

\subheader{Detectors} 
We evaluated the performance of a broad set of detection algorithms. These include both publicly available pretrained detectors made specifically for synthetic image detection, and other architectures that are widely used to perform image forensic tasks. The complete list of these detectors and their referred-to names is provided in Table~\ref{tab:list_of_detectors}.

\subheader{Image Training Dataset} 
To train detectors that weren't pretrained, we used a dataset of 100,000 images equally divided between real and synthetic.  For real images, we utilized a subset of the COCO dataset~\cite{coco} and the LSUN dataset~\cite{lsun}, as was done in~\cite{luisa}.  For synthetic images, we used synthetic images from the datasets used in~\cite{openset} consisting of images made by CycleGAN~\cite{cyclegan}, StarGAN~\cite{stargan}, StyleGAN3~\cite{stylegan3}, ProGAN~\cite{progan}, and Stable Diffusion~\cite{stablediffusion}.

\subheader{Image Testing Dataset}
To benchmark the performance of each detector, we created a testing set of 20,000 images equally divided between real and synthetic.  This set was made by utilizing disjoint subsets of the datasets used to create the real and synthetic image training data.

\subheader{Video Testing Dataset}
To measure the performance of each detector, we created a testing dataset of both real and synthetically generated videos. 
Real videos were taken equally from the Moments in Time (MiT)~\cite{mit} and Video-ACID~\cite{videoacid} datasets.
Synthetic videos were generated using four different publicly available video generators: Luma~\cite{Luma}, VideoCrafter-v1~\cite{videocrafter1}, CogVideo~\cite{cogvideo}, and Stable Video Diffusion~\cite{stablevideo}.  
These synthetic videos were created using a common set of diversified content and motion text prompts, with the exception of videos from Luma, which were gathered from a similarly diverse set of publicly shared videos. 
The qualitative samples of these videos are shown in Fig.~\ref{fig:video_gen_visualization}. Further details of this test set are provided in Table~\ref{tab:video_dataset_stats} and in Sec.~\ref{subsec:common_video_exp_setup} below.

\subheader{Metrics}  The detection performance of each detector was measured using the area under its ROC curve (AUC).

\subsection{Synthetic Image Detector Performance}

\begin{table}[!t]
	\centering
	\resizebox{\linewidth}{!}{%
		\begin{tblr}{
				width = \linewidth,
				colspec = {m{24mm}m{12mm}m{9mm}m{14mm}m{10mm}m{8mm}m{12mm}},
				row{1-2} = {c},
				column{2-8} = {c},
				column{1} = {l},
				cell{1}{1} = {r=2}{},
				cell{1}{3} = {c=5}{0.634\linewidth,c},
				vline{1-3, 7-8} = {-}{},
				hline{1-3, 6,9, 14} = {-}{},
			}
			\textbf{Method} & \textbf{Images} & \textbf{Videos} &  &  &  & \\
			& \textbf{Baseline} & \textbf{Luma} & \textbf{CogVideo} & \textbf{VC-v1} & \textbf{SVD} & \textbf{Average}\\
			Corvi et al.~\cite{luisa} & 0.974 & 0.583 & 0.704 & 0.590 & 0.682 & 0.640 \\
			Sinitsa et al.~\cite{dif} & 0.992 & 0.500 & 0.500 & 0.500 & 0.500 & 0.500 \\
			Zhu et al.~\cite{genimage} & 0.891 & 0.652 & 0.694 & 0.728 & 0.719 & 0.698 \\
			ResNet-50~\cite{resnet_archit} & 0.946 & 0.572 & 0.736 & 0.604 & 0.710 & 0.656 \\
			DIF~\cite{dif} & 0.991 & 0.581 & 0.603 & 0.617 & 0.573 & 0.594 \\
			Swin-T~\cite{swint} & 0.911 & 0.638 & 0.685 & 0.698 & 0.692 & 0.678 \\
			ResNet-34~\cite{resnet_archit} & 0.983 & 0.576 & 0.623 & 0.616 & 0.647 & 0.615 \\
			VGG-16~\cite{vgg16_archit} & 0.990 & 0.635 & 0.652 & 0.684 & 0.669 & {0.660} \\
			Xception~\cite{xception_archit} & 0.996 & 0.592 & 0.638 & 0.670 & 0.664 & 0.641 \\
			DenseNet~\cite{densenet_archit} & 0.975 & 0.559 & 0.584 & 0.647 & 0.678 & 0.624 \\
            MISLnet~\cite{openset} & 0.983 & 0.626 & 0.718 & 0.710 & 0.707 & 0.690 \\
		\end{tblr}
	}
    \caption{Detection performance of existing synthetic image detectors, that were trained or pretrained on synthetic images, on different synthetic video generation methods. Performance numbers are measured using AUC.}
    \label{tab:synth_im_det_on_videos}
    \pulluppp
\end{table}

We first established the baseline performance of each synthetic image detector on our image testing dataset.
These results are presented in the second column of Table~\ref{tab:synth_im_det_on_videos}. The majority of detectors achieved an AUC of 0.94 or greater, except for the pre-trained version of Swin-T with an AUC of 0.891.  
These baseline results verify that when assessed on images, each detector can achieve strong performance.

Next we evaluated each synthetic image detector using our video testing dataset.  These results are also shown in Table~\ref{tab:synth_im_det_on_videos}.
These results show that all detectors experience significant performance drops when evaluating synthetic videos. The highest average AUC achieved was 0.698, with most detectors scoring an AUC of 0.65 or lower. This drop in performance cannot be attributed to a single challenging generator, as AUCs for each detector on a single generator are consistently less than 0.74.

These results demonstrate that synthetic image detectors face significant challenges in detecting synthetic videos. This difficulty persists across various detector architectures and whether detectors are pre-trained by others or retrained using our dataset.

\subsection{Effect of H.264 Robust Training}

\begin{table}[!t]
	\centering
	\resizebox{\linewidth}{!}{%
		\begin{tblr}{
				width = \linewidth,
				colspec = {m{22mm}m{12mm}m{8mm}m{13mm}m{10mm}m{8mm}m{8mm}m{14mm}},
				row{1-2} = {c},
				column{2-9} = {c},
				column{1} = {l},
				cell{1}{1} = {r=2}{},
				cell{1}{3} = {c=6}{0.634\linewidth,c},
				vline{1-3, 7-9} = {-}{},
				hline{1-3,6, 11} = {-}{},
			}
			\textbf{Method} & \textbf{Images} & \textbf{Videos} &  &  &  & \\
			& \textbf{Baseline} & \textbf{Luma} & \textbf{CogVideo} & \textbf{VC-v1} & \textbf{SVD} & \textbf{Avg.}& {\textbf{Vs.}\\\textbf{no H.264}}\\
			ResNet-50~\cite{resnet_archit} & 0.963 & 0.604 & 0.770 & 0.646 & 0.738 & 0.689 & +0.033 \\
			DIF~\cite{dif} & 0.994 & 0.617 & 0.634 & 0.655 & 0.624 & 0.632 & +0.038 \\
			Swin-T~\cite{swint} & 0.948 & 0.679 & 0.730 & 0.758 & 0.742 & 0.727& +0.049 \\
            ResNet-34~\cite{resnet_archit} & 0.989 & 0.663 & 0.687 & 0.700 & 0.727 & 0.694 & +0.079 \\
            VGG-16~\cite{vgg16_archit} & 0.993 & 0.719 & 0.743 & 0.754 & 0.729 & 0.736 & +0.076 \\
            Xception~\cite{xception_archit} & 0.979 & 0.642 & 0.692 & 0.734 & 0.708 & 0.694 & +0.053 \\
            DenseNet ~\cite{densenet_archit} & 0.980 & 0.604 & 0.628 & 0.691 & 0.703 & 0.656 & +0.032 \\
            MISLnet ~\cite{openset} & 0.995 & 0.674 & 0.759 & 0.784 & 0.760 & 0.744 & +0.054 \\
		\end{tblr}
	}
    \caption{Detection performance of existing synthetic image detectors, that were retrained on H.264-compressed synthetic images, on different synthetic video generation methods. Performance numbers are measured using AUC.}
    \label{tab:synth_im_det_on_videos_h264}
    \pulluppp
\end{table}

\begin{figure*}[!t]
	\centering
	
	\begin{minipage}[t]{1.0\textwidth}
		\centering
		\setlength{\fboxrule}{0pt}
        \setlength{\fboxsep}{0.5pt}
		\fbox{\parbox{0.150\textwidth}{\centering\smaller ProGAN}}
		\fbox{\parbox{0.150\textwidth}{\centering\smaller Latent Diffusion}}
		\fbox{\parbox{0.150\textwidth}{\centering\smaller CycleGAN}}
		\fbox{\parbox{0.150\textwidth}{\centering\smaller StarGAN}}
		\fbox{\parbox{0.150\textwidth}{\centering\smaller StyleGAN 3}}
		\fbox{\parbox{0.150\textwidth}{\centering\smaller Stable Image Diffusion}}
		\smallskip
	\end{minipage}
	
	\begin{minipage}[t]{1.0\textwidth}
		\centering
		\setlength{\fboxrule}{0pt}
        \setlength{\fboxsep}{0.5pt}
		\fbox{\includegraphics[width=0.150\textwidth]{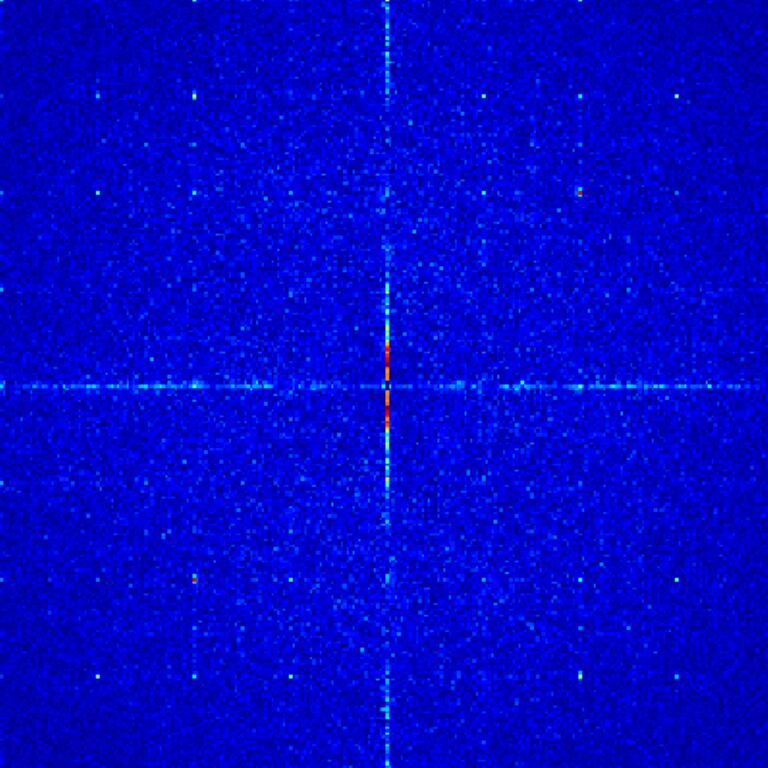}}
		\fbox{\includegraphics[width=0.150\textwidth]{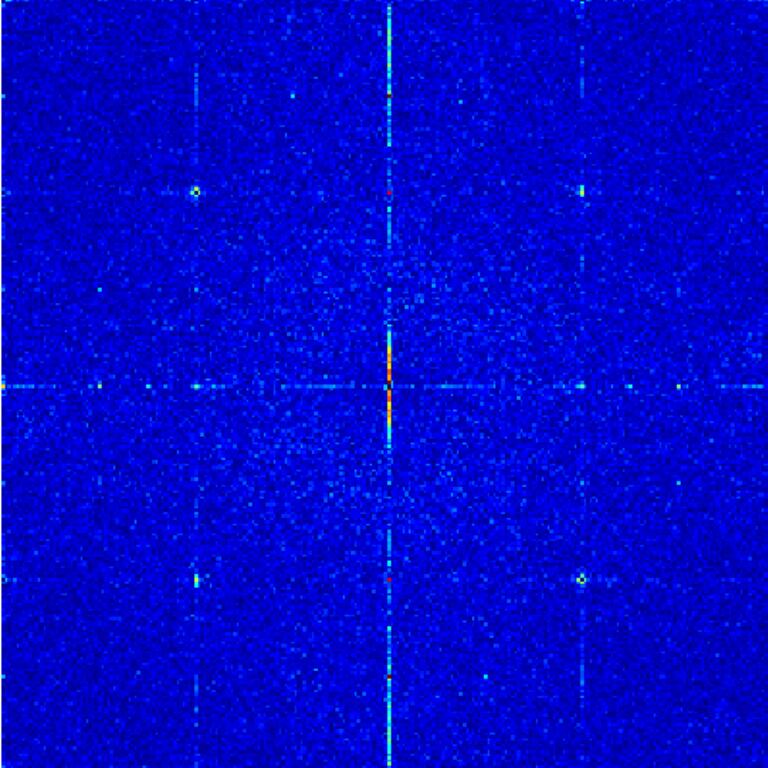}}
		\fbox{\includegraphics[width=0.150\textwidth]{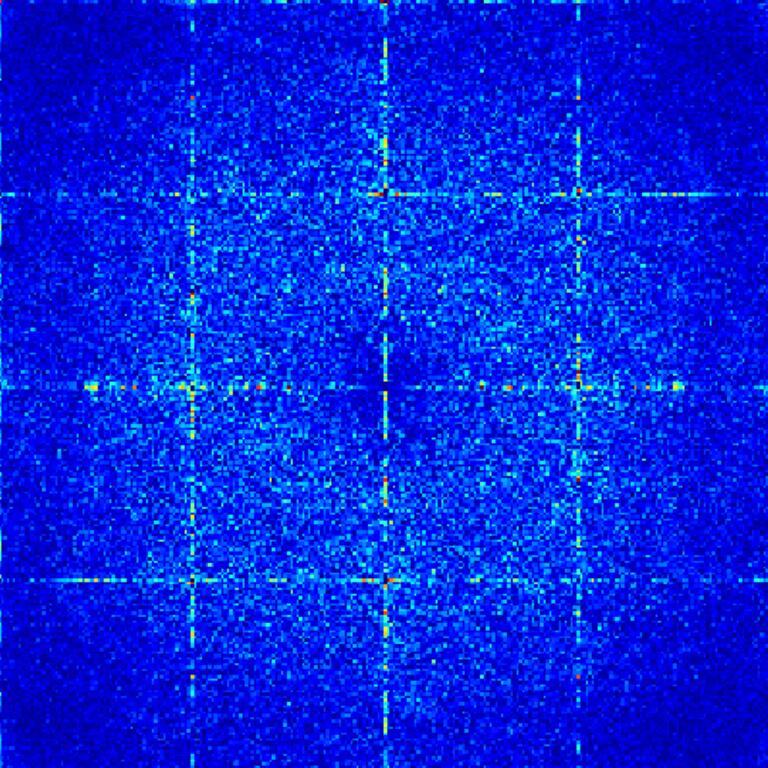}}
		\fbox{\includegraphics[width=0.150\textwidth]{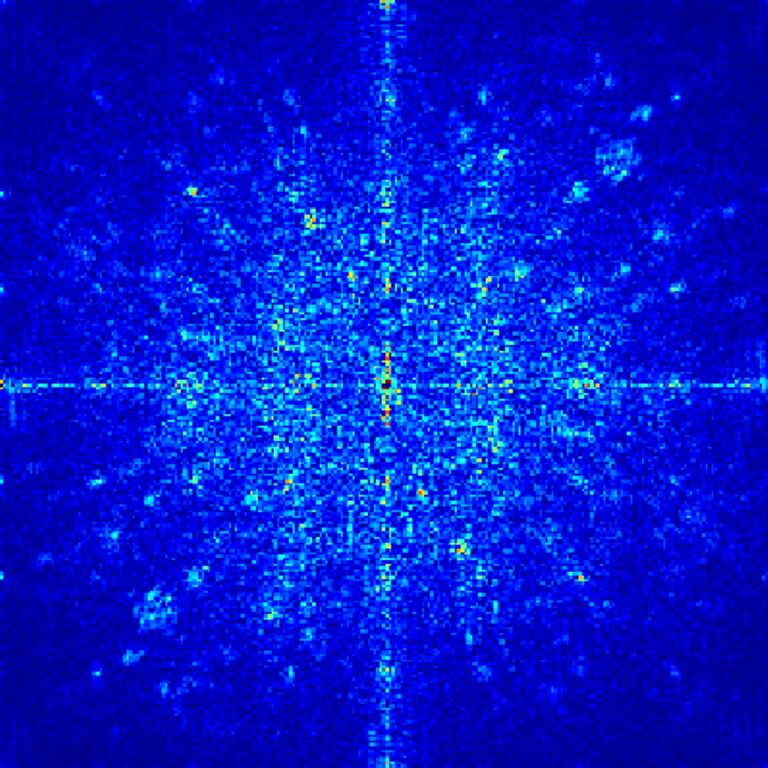}}
		\fbox{\includegraphics[width=0.150\textwidth]{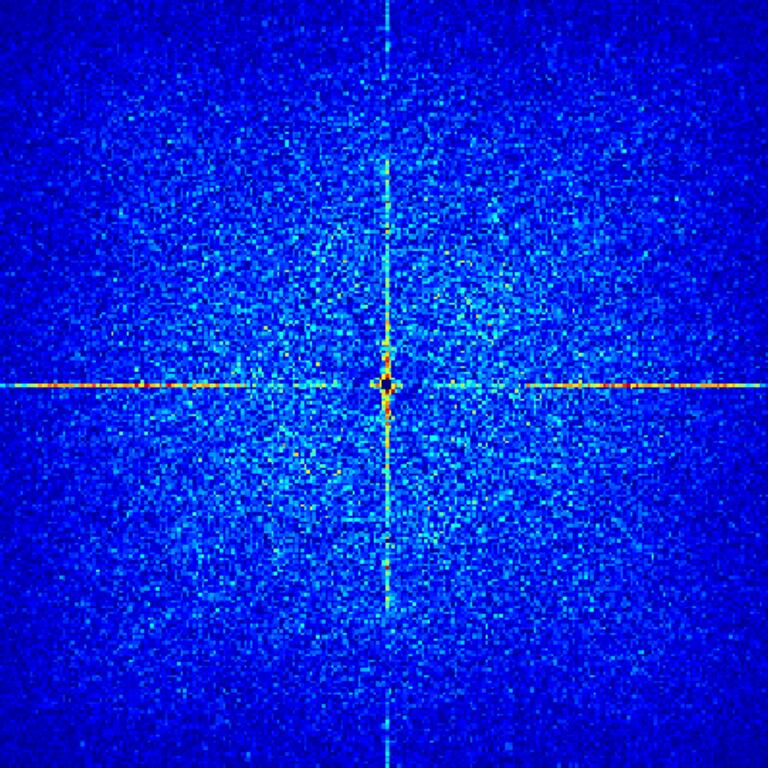}}
		\fbox{\includegraphics[width=0.150\textwidth]{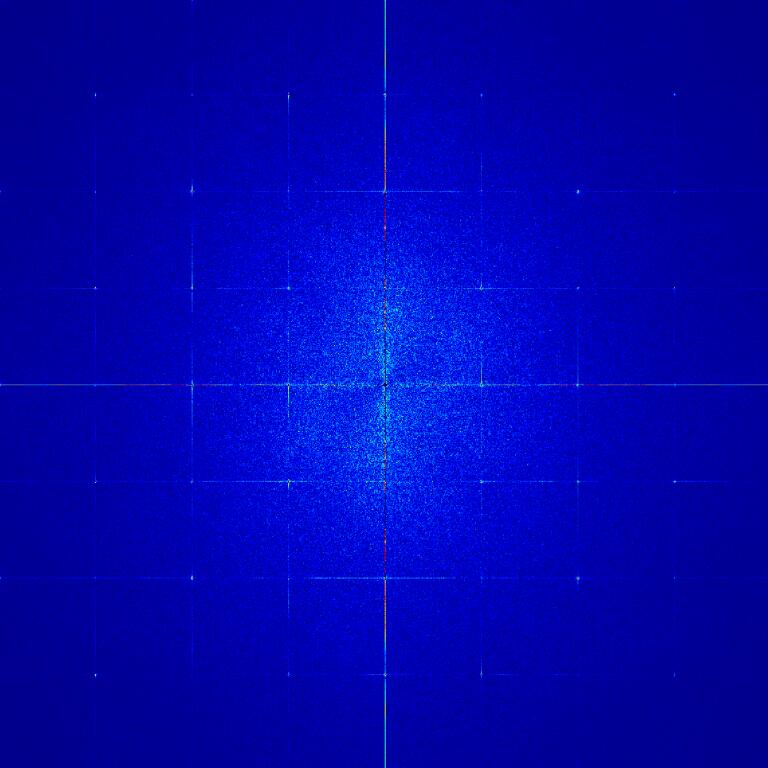}}
	\end{minipage}
	
	\begin{minipage}[t]{1.0\textwidth}
		\centering
		\setlength{\fboxrule}{0pt}
        \setlength{\fboxsep}{0.5pt}
		\fbox{\parbox{0.150\textwidth}{\centering\smaller Luma}}
		\fbox{\parbox{0.150\textwidth}{\centering\smaller VideoCrafter}}
		\fbox{\parbox{0.150\textwidth}{\centering\smaller CogVideo}}
		\fbox{\parbox{0.150\textwidth}{\centering\smaller Pika}}
		\fbox{\parbox{0.150\textwidth}{\centering\smaller Sora}}
		\fbox{\parbox{0.150\textwidth}{\centering\smaller Stable Video Diffusion}}
		\smallskip
	\end{minipage}
	
	\begin{minipage}[t]{1.0\textwidth}
		\centering
		\setlength{\fboxrule}{0pt}
        \setlength{\fboxsep}{0.5pt}
		\fbox{\includegraphics[width=0.150\textwidth]{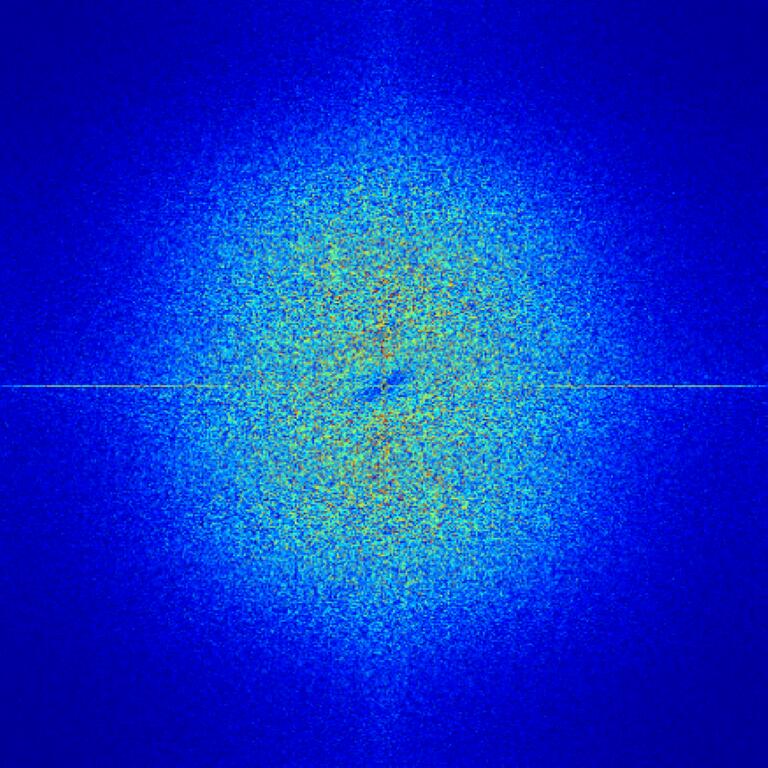}}
		\fbox{\includegraphics[width=0.150\textwidth]{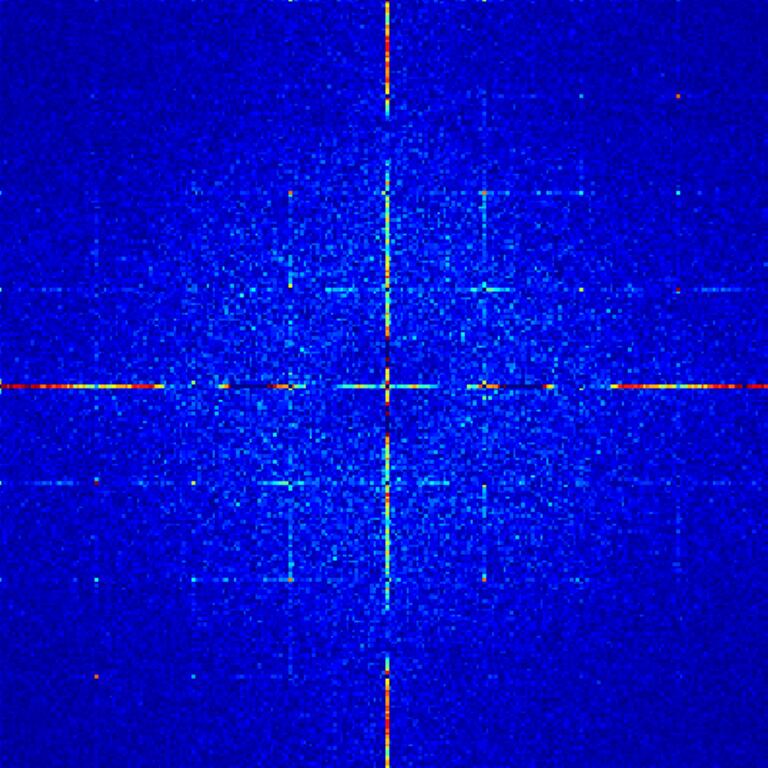}}
		\fbox{\includegraphics[width=0.150\textwidth]{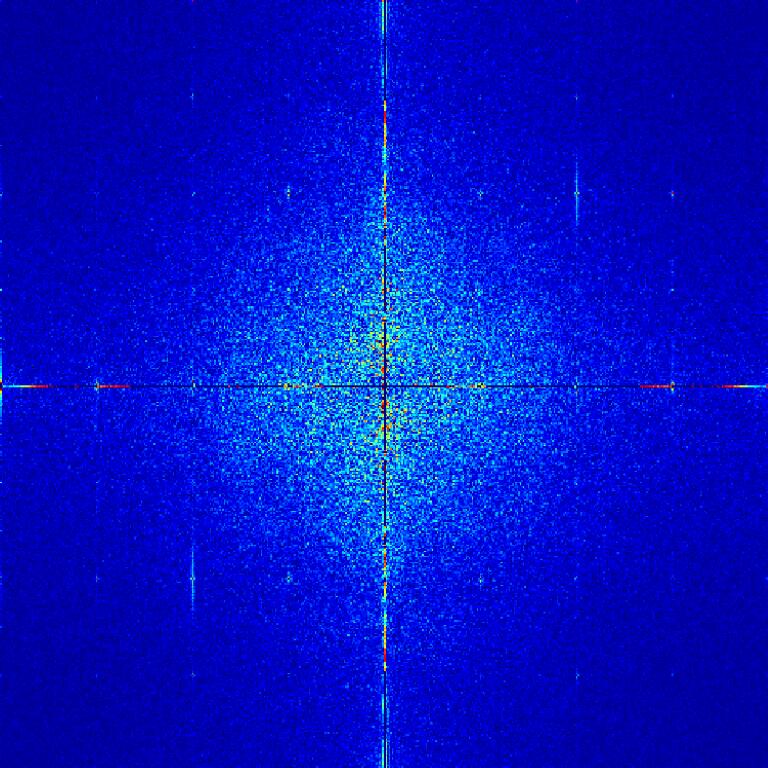}}
		\fbox{\includegraphics[width=0.150\textwidth]{{figures/trace_visualization/pika}}}
		\fbox{\includegraphics[width=0.150\textwidth]{{figures/trace_visualization/sora}}}
		\fbox{\includegraphics[width=0.150\textwidth]{{figures/trace_visualization/stable_vid}}}
	\end{minipage}
	
	\caption{Fourier transform analysis of the forensic traces extracted from different synthetic image and video generators.}
	\label{fig:trace_visualization}
    \pulluppp\pullup
\end{figure*}

It is well known that compression alters forensic traces and degrade a detector's performance.  
Hence, a plausible explanation for synthetic image detector's the poor performance on synthetic video could be that H.264 video compression is degrading synthetic video traces\cite{nguyen2024videofact}.

To test this hypothesis, we conducted an additional set of experiments in which each synthetic image detector was retrained to be robust against H.264 compression.  
Robust training involves augmenting the training dataset by re-compressing all data points using different compression strengths, i.e. different quality factor for JPEG or different compression rate factors (CRFs) for H.264. 
It is well known that robust training can significantly mitigate the negative effects of compression~\cite{luisa,compression1,compression2,compression3}.  As a result, if H.264 compression was truly the cause of a detector's low performance on synthetic video, we would expect the detector's performance to increase to a level much closer to its baseline performance after robust training.

Table~\ref{tab:synth_im_det_on_videos_h264} shows the performance of each synthetic image detector after it was robustly trained by augmenting the image training dataset with  H.264 compressed images using CRFs between 0 and 20.
The baseline detector performances on images show that robust training improves each detector's already strong performance on images.
Despite this, robust training did not substantially improve each detector's performance when evaluated on our video testing set. The detector with the strongest performance on video after robust training was VGG-16, which obtained an AUC of 0.736 on video as opposed to its baseline performance of 0.993 on images.  
Most detectors achieved an average AUC of below 0.74. And on average, robust training only improved the AUC of each detector by 0.052. The largest AUC gain was seen by ResNet-34~\cite{resnet34_method,resnet_archit}, increasing its average AUC by 0.079 to 0.694.

These results indicate that H.264 is not the primary cause of synthetic image detectors' poor performance when detecting synthetic videos.  Instead, the poor performance of synthetic image detectors after H.264 robust training suggests that a different factor is causing this phenomenon.



\section{Synthetic Video Forensic Traces}
\label{sec:synth_video_traces}

Here, we present evidence that forensic traces in synthetic video are substantially different than those in synthetic images.  We qualitatively demonstrate this by visualizing the low-level forensic traces left by a number of different image and video generators using the approach proposed in~\cite{resnet34_method}. 

To do this, we collected a set of 1000 images and video frames created using several different image and video generators.  These image generators included ProGAN~\cite{progan}, CycleGAN~\cite{cyclegan}, StarGAN~\cite{stargan}, StyleGAN 3~\cite{stylegan3}, Latent Diffusion~\cite{diffusion7}, and Stable Diffusion~\cite{stablediffusion}, while video generators included Luma~\cite{Luma}, VideoCrafter v1~\cite{videocrafter1}, CogVideo~\cite{cogvideo}, Pika~\cite{Pika}, Sora~\cite{sora}, and Stable Video Diffusion~\cite{stablevideo}. 
We then created a noise residual for each image and video frame $x_k$ by de-noising it using a de-noising algorithm $\phi$, then subtracting the denoised image or frame from the original.  
All noise residuals from a single generator were averaged to produce an aggregate noise residual $y= \tfrac{1}{N}\sum_{k=1}^{N}x_k-\phi(x_k)$.  Frequency domain representations of these aggregate noise residuals were then created by taking their Fourier transforms, then their magnitudes were plotted to produce trace visualizations.

\subheader{Image and Video Generator's Trace Comparison}
The resulting low-level forensic traces for each image and video generator are shown in Fig.~\ref{fig:trace_visualization}.  By examining these traces, we can clearly see that synthetic images contain substantially different traces than synthetic videos.
For example, traces left by image generators (e.g. ProGAN, CycleGAN, Latent Diffusion) typically include periodic spectral peaks or a grid-like structure that Zhang et al.~\cite{resnet34_method} showed are caused by up-sampling operations used in a generator's architecture.  By contrast, some video generation techniques such as Luma employ neural radiance fields (NeRFs) or other architectures that do not utilize up-sampling.  As a result, the distinct patterns seen in synthetic image traces will not be present in traces from these video generators.

Additionally, industry generators may use undisclosed techniques to protect trade secrets. For instance, full details about Pika's generation method are not currently public, but the significant difference between its traces and others' suggests Pika uses a noticeably different technique.

Due to the stark contrasts between forensic traces left by image and video generators, it is highly likely that this is the major reason why synthetic image detectors exhibit substantially lower performance on video.  Even when robustly trained, synthetic image detectors learn features to capture forensic traces similar to what they have seen before.  Since video traces can be substantially different in nature, synthetic image detectors are not suited to capture these traces.  

We note that our findings align with prior research. Specifically, Corvi et al.~\cite{luisa} found that Stable and Latent diffusion models produce different forensic traces than image generators such as ADM and DALL-E 2.  They also showed that even robustly trained synthetic image detectors ``still cannot reliably detect images that present artifacts significantly different from those seen during training.''~\cite{luisa}



\section{Learning Synthetic Video Forensic Traces}
\label{sec:learn_synth_vid_traces}

Results presented in the previous two sections show that traces left by synthetic video generators are different than those left by image generators, and that synthetic image detectors do not reliably detect these traces.

In this section, however, we show that synthetic video traces can be learned.  Through a series of experiments, we show that CNNs can be trained to accurately perform synthetic video detection and source attribution.   
Furthermore, we demonstrate that robust training can improve these detectors even after H.264 re-compression.  Additionally, we show how video-level detection can be performed to boost performance over frame-level detection.

\subsection{Experimental Setup}
\label{subsec:common_video_exp_setup}

\definecolor{Mercury}{rgb}{0.913,0.913,0.913}
\begin{table}
    \centering
    \resizebox{\linewidth}{!}{%
    \begin{tblr}{
      width = \linewidth,
      colspec = {m{3mm}m{25mm}m{13mm}m{14mm}m{13mm}m{14mm}m{13mm}m{14mm}},
      row{1,2} = {c},
      column{3-8} = {c},
      column{2} = {l},
      row{5, 10,14} = {Mercury},
      cell{1}{1} = {r=2}{},
      cell{1}{2} = {r=2}{},
      cell{1}{3} = {c=2}{0.244\linewidth},
      cell{1}{5} = {c=2}{0.244\linewidth},
      cell{1}{7} = {c=2}{0.244\linewidth},
      cell{3}{1} = {r=2}{c},
      cell{5}{1} = {c=2}{0.192\linewidth},
      cell{6}{1} = {r=4}{c},
      cell{10}{1} = {c=2}{0.192\linewidth},
      vline{1,9} = {-}{},
      vline{2-3,5,7} = {1-4,5-10}{},
      hline{1,3,5-6,10-11} = {-}{},
      hline{2} = {3-8}{},
    }
        & \textbf{Dataset} & \textbf{Training} &  & \textbf{Validation} &  & \textbf{Testing} & \\
        &  & \textbf{\# Videos} & \textbf{\# Frames} & \textbf{\# Videos} & \textbf{\# Frames} & \textbf{\# Videos} & \textbf{\# Frames}\\
        \begin{sideways}\textbf{Real}\end{sideways}
        & MIT~\cite{mit} & 3,991  & 80,000  & 377 & 8,000 & 945 & 20,000 \\
        & Video-ACID~\cite{videoacid} & 3,663  & 80,000 & 407 & 8,000 & 716 & 20,000 \\
        \textbf{Total} &  & 7,654 & 160,000 & 784 & 16,000 & 1,661 & 40,000 \\
        
        \begin{sideways}\textbf{Synthetic}\end{sideways}
        & Luma~\cite{Luma} & 312 & 40,000 & 32 & 4,000 & 78 & 10,000 \\
        & VC-v1~\cite{videocrafter1} & 1,428 & 40,000 & 143 & 4,000 & 280 & 10,000 \\
        & CogVideo~\cite{cogvideo} & 1,600 & 40,000 & 163 & 4,000 & 357 & 10,000 \\
        & SVD~\cite{stablevideo} & 2,857 & 40,000 & 286 & 4,000 & 714 & 10,000\\
        \textbf{Total} &  & 6,197 & 160,000 & 624 & 16,000 & 1,429 & 40,000  \\

    \end{tblr}
    }
    \caption{Dataset statistics for training and evaluating detection systems on synthetic video data. VC stands for VideoCrafter.}
    \label{tab:video_dataset_stats}
    \pulluppp\pullupp
\end{table}

The following experiments all used the same experimental setup detailed here. 

\subheader{Video Training Data}
To train synthetic video detectors in the following experiments, we collected a diverse set of real and synthetic videos. 
For real videos, we gathered videos from the Moments in Time (MIT)~\cite{mit} dataset and the Video Authentication and Camera Identification Database (Video-ACID)~\cite{videoacid}.
For the set of synthetic videos, we used 4 publicly available video generators to generate a large dataset for both training and testing purposes. These generation methods are: Luma~\cite{Luma}, VideoCrafter-v1~\cite{videocrafter1}, CogVideo~\cite{cogvideo}, and Stable Video Diffusion~\cite{stablevideo}. 
To create the synthetic videos in our dataset, we utilized a common set of text prompts chosen to represent a diversified set of scenes and activities.  Videos created using Luma were gathered from publicly shared videos, and similarly chosen to represent a diverse set of contents and motion. 
Additionally, all videos are by default compressed using H.264 at constant rate factor 23. More detailed are provided in Table~\ref{tab:video_dataset_stats}.

\subheader{Video Testing Data}
To create our video testing data, we utilized the same collection and generation methodology as for our training data. However, we kept an exclusive set of testing prompts to only be used to create testing data. Regarding videos from Luma, we also gathered a disjoint set of videos, completely separate from those in the training set. We note that this testing set is the same set used for experiments in Sec.~\ref{sec:synth_img_det_on_videos}. More detailed are provided in Table~\ref{tab:video_dataset_stats}.

\subheader{Out-of-distribution Test-only Synthetic Videos}
In addition to our in-distribution video testing set, we also evaluated the performance of different detection algorithms on an out-of-distribution, test-only set of videos. The synthetic videos in this set is collected from three recently emerging generation methods: Sora~\cite{sora}, Pika~\cite{Pika}, and VideoCrafter-v2~\cite{videocrafter2}. More details on this dataset are provided in Table~\ref{tab:ood_dataset_stats}

\subheader{Metrics}
In the following experiments, the performance of each detector was measured using the area under its ROC curve (AUC). In addition, to highlight performance differences, we provided the Relative Error Reduction (RER) with respect to the second best performing method, reported as a percentage. This metric is calculated as follows:
\skipless
\begin{equation}
	RER = 100 \times \frac{\text{AUC}_{N} - \text{AUC}_{R}}{1 - \text{AUC}_{R}}, 
\end{equation}
where $\text{AUC}_{R}$ is the AUC of the referencing method, and $\text{AUC}_{N}$ is the AUC of the method being compared against.

\skipnormal

\subheader{Detectors}
To demonstrate the performance of different detection methods on detecting synthetic videos, we conducted our experiments with the diverse set of detectors listed in Table~\ref{tab:list_of_detectors} and Sec.~\ref{subsec:synth_img_exp_setup}.

\subsection{Synthetic Video Detection}
\label{subsec:exp1_synth_vid_det}

First, we conducted an experiment in which we trained each candidate detector network to perform synthetic video detection using the training dataset described above. We then evaluated their ability to detect each of the four synthetic video generators in the test set.  These experiments were carried out at a patch-level, i.e. all detection decisions were obtained using one patch taken from a single video frame.

\begin{table}[!t]
	\centering
	\resizebox{\linewidth}{!}{%
		\begin{tblr}{
				width = \linewidth,
				colspec = {m{26mm}m{12mm}m{16mm}m{14mm}m{9mm}m{14mm}},
				row{1-2} = {c},
				column{2-8} = {c},
				column{1} = {l},
				cell{1}{1} = {r=2}{},
				cell{1}{2} = {c=5}{0.634\linewidth,c},
				vline{1-2, 6, 7} = {-}{},
				hline{1-3,6, 11} = {-}{},
			}
			\textbf{Method}   & \textbf{Videos} &  &  &  & \\
			  & \textbf{Luma} & \textbf{CogVideo} & \textbf{VC-v1} & \textbf{SVD} & \textbf{Average}\\
			ResNet-50~\cite{resnet_archit}  & 0.921 & \textbf{0.935} & 0.940 & 0.939 & 0.934 \\
			DIF~\cite{dif}  & 0.938 & 0.957 & 0.969 & 0.973 & 0.959 \\
			Swin-T~\cite{swint}  & 0.960 & 0.964 & 0.986 & 0.991 & 0.975 \\
            ResNet-34~\cite{resnet_archit}  & 0.916 & 0.930 & 0.942 & 0.951 & 0.935  \\
            VGG-16~\cite{vgg16_archit}  & 0.937 & 0.944 & 0.965 & 0.960 & 0.951  \\
            Xception~\cite{xception_archit}  & 0.928 & 0.951 & 0.973 & 0.969 & 0.955  \\
            DenseNet ~\cite{densenet_archit}  & 0.918 & 0.924 & 0.964 & 0.968 & 0.943  \\
            MISLnet ~\cite{openset}  & \textbf{0.975} & \textbf{0.980} & \textbf{0.991} & \textbf{0.987} & \textbf{0.983}  \\
		\end{tblr}
	}
    \caption{Detection performance of methods trained and tested on different synthetic video generation methods. Performance numbers are measured using AUC.}
    \label{tab:exp1_synth_vid_det}
    \pullupp\pullup
\end{table}

The results of this experiment are presented in Table~\ref{tab:exp1_synth_vid_det}.  These results clearly show that synthetic videos can be reliably detected using each of these detectors.  All detectors evaluated achieved an average AUC of at least 0.93.  MISLnet achieved the highest average AUC of 0.983 and maintained consistently strong detection performance for each video generator.  
We note that each detector trained on synthetic video experienced an improvement of at least 0.23 in average AUC when compared to performance of the same detector robustly trained on synthetic images.
This further reinforces that synthetic video traces can be learned by existing architectures used for synthetic image detection.

\subsection{Synthetic Video Source Attribution}
\label{subsec:exp2_synth_vid_attr}

Next, we conducted an experiment evaluating each network's ability in source attribution. The forensic network identifies a video's source generator or determines its authenticity. To adapt each network for this multi-class classification, we replaced its final layer with one containing neurons corresponding to each generator and one for real. The trained network's AUC was assessed using the one-vs-the-rest strategy, where any incorrect source attribution was counted as a miss regardless of the class.

\begin{table}[!t]
	\centering
	\resizebox{\linewidth}{!}{%
		\begin{tblr}{
				width = \linewidth,
				colspec = {m{26mm}m{12mm}m{16mm}m{14mm}m{9mm}m{14mm}},
				row{1-2} = {c},
				column{2-8} = {c},
				column{1} = {l},
				cell{1}{1} = {r=2}{},
				cell{1}{2} = {c=5}{0.634\linewidth,c},
				vline{1-2, 6, 7} = {-}{},
				hline{1-3,6, 11} = {-}{},
			}
			\textbf{Method}   & \textbf{Videos} &  &  &  & \\
			  & \textbf{Luma} & \textbf{CogVideo} & \textbf{VC-v1} & \textbf{SVD} & \textbf{Overall}\\
			Resnet-50~\cite{luisa}  & 0.937 & 0.958 & 0.970 & 0.968 & 0.962 \\
			DIF~\cite{dif}  & 0.894 & 0.909 & 0.926 & 0.924 & 0.917 \\
			Swin-T~\cite{swint}  & \textbf{0.976} & 0.973 & 0.991 & 0.988 & 0.986 \\
            ResNet-34~\cite{resnet_archit}  & 0.925 & 0.947 & 0.963 & 0.932 & 0.948  \\
            VGG-16~\cite{vgg16_archit}  & 0.901 & 0.936 & 0.945 & 0.959 & 0.935  \\
            Xception~\cite{xception_archit}  & 0.925 & 0.971 & 0.960 & 0.952 & 0.950  \\
            DenseNet~\cite{densenet_archit}  & 0.954 & 0.959 & 0.976 & 0.970 & 0.966  \\
            MISLnet~\cite{openset}  & 0.975 & \textbf{0.984} & \textbf{0.998} & \textbf{0.992} & \textbf{0.991}  \\
		\end{tblr}
	}
    \caption{Synthetic video source attribution performance of each detection systems on individual generation method. Performance numbers are measured using AUC.}
    \label{tab:exp2_synth_vid_attr}
    \pulluppp\pullupp
\end{table}

The results of this experiment are shown in Table \ref{tab:exp2_synth_vid_attr}.  From these results, we can see that 
all networks achieved an AUC of at least 0.91, with most achieving an AUC of 0.95 or higher.  Again, the best performing network was MISLnet, which acheived an AUC of 0.991.
These results indicate that existing networks can be trained to perform accurate synthetic video source attribution.  We note that this result makes sense in light of the synthetic video traces visualized in Fig.~\ref{fig:trace_visualization}.  As discussed in Sec.~\ref{sec:synth_video_traces}, videos are generated using a wide variety of generation strategies and generator architectures. Since each technique imparts significant different traces, this makes it much easier for networks to accurately discriminate between each generator.

\subsection{Effect of H.264 Re-compression}
\label{subsec:exp3_vid_compression}

\begin{figure}[!t]
    \centering
    \includegraphics[width=6.5cm,keepaspectratio]{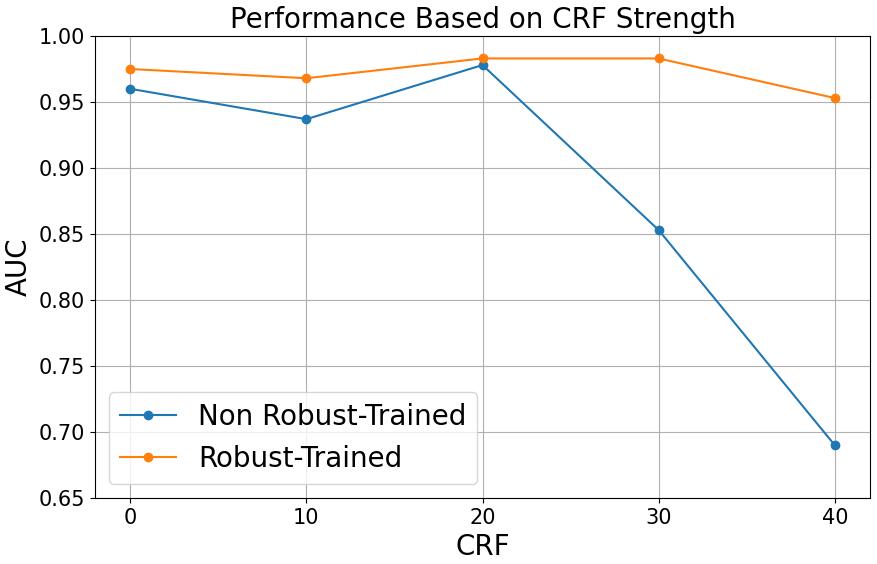}
    \caption{Detection performance of MISLnet\cite{openset} before and after robust-training on videos with constant rate factors from 0 to 40.}
    \label{fig:exp3_recompression}
\end{figure}

As we discussed in Sec.~\ref{sec:synth_img_det_on_videos}, it is well known that re-compression can significantly reduce the performance of a forensic system.  
This is particularly important since re-compression is often utilized by social media platforms.  
In light of this, we conducted a set of experiments to understand the effect of H.264 compression on detection performance.  Additionally, we conducted experiments to assess the ability of robust training to mitigate these effects.
In light of space constraints, results are reported for the MISLnet detector, which achieved the highest detection and attribution performance in the previous experiments.

To carry out these experiments, we re-compressed each video in the testing set with constant rate factors (CRFs) ranging from 0 (weak) to 40 (strong). We note that videos are all initially H.264 compressed at CRF 23 either by the camera or the generator.
We then re-compressed each video in the training set using the same CRF levels, and used them to robustly train MISLnet to perform synthetic video detection.
After this, we evaluated the performance of both the non-robustly and robustly trained version of this detector.

The results of this experiment are displayed in Fig.~\ref{fig:exp3_recompression}, which shows the AUC achieved by the detector at each CRF both with and without robust training.  
From these results, we can see that without robust training, the detector's performance decreases as the CRF increases.  The notable exception to this is at CRF 20, which is close to the typical default CRF of 23.  Performance increases here because the CRF used during re-compression is close to the default CRF that has already been seen during training.  

The results presented in Fig.~\ref{fig:exp3_recompression} also show that when robust training is utilized, the detector's performance remains consistently strong across all CRFs.
Specifically, the detector is able to achieve an AUC of 0.95 or higher for all CRFs.  Furthermore, when the CRF is 30 or higher, the detector achieves an AUC of 0.97 or higher. These results demonstrate that robust training enables accurate synthetic video detection even after re-compression.

\subsection{Video-Level Detection Performance}
\label{subsec:exp4_vid_level_analysis}

Unlike synthetic images, synthetic videos consist of a sequence of AI-generated frames.
Because of this, generator traces are distributed temporally throughout a video.  This information can be exploited to perform detection at a video-level with greater accuracy.

\begin{figure}[!t]
    \centering
    \includegraphics[width=6.5cm,keepaspectratio]{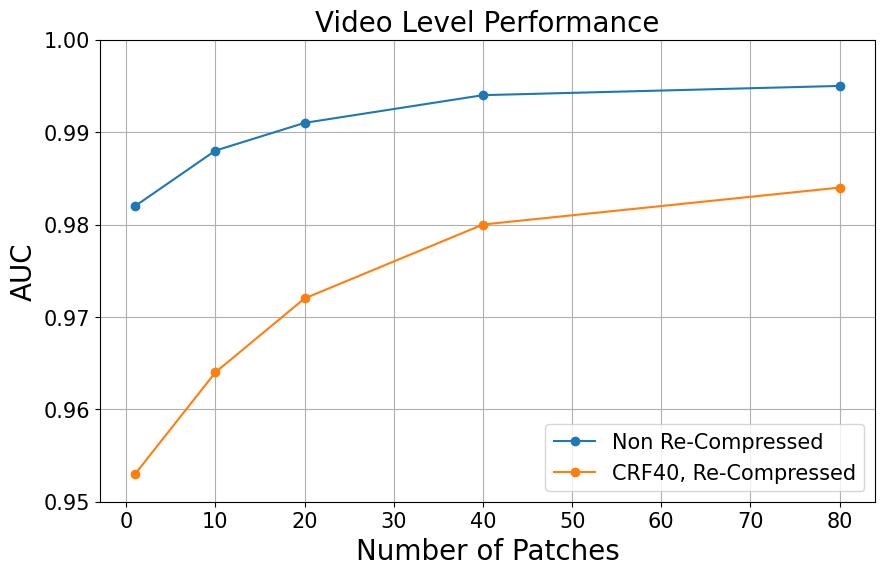}
    \caption{Video-level performance of MISLnet over different number of patches used for obtaining video-level detection score.}
    \label{fig:video_auc_plot}
    \pulluppp\pullupp
\end{figure}

To perform video-level detection, we first form a patch-level embedder $\psi(\cdot)$ by discarding the final layer of a pre-trained patch-level synthetic video detector.  The pre-softmax activations produced by this network correspond to patch-level embeddings that capture video generator traces.
Next, a sequence of $N$ temporally distributed patches $x_k$ are gathered throughout a video and added together to form a single video-level embedding.  This is then passed through a soft-max layer to produce the final output detection score 
\skipless
\begin{equation}
    \delta= \sigma\Bigl( \sum\nolimits_{k=1}^N \psi(x_k) \Bigr),
\end{equation}
where $\sigma(\cdot)$ is the soft-max function\cite{hosler2019video,mayer2020open}.

We conducted a series of experiments where we measured the performance of this video-level detection strategy using different numbers of patches.  In these experiments, we used the MISLnet detector architecture since it achieved the best performance for patch-level detection and attribution.
Performance in terms of AUC was measured on both our non-recompressed testing set, as well as on a version of the testing set that was re-compressed with a CRF of 40 for different numbers of patches $N$ ranging from 1 to 80.  Additionally, the percentage of relative error reduction (RER) over a patch-level detector was also calculated for each $N$.

\begin{figure}[!t]
    \centering
    \includegraphics[width=6.5cm,keepaspectratio]{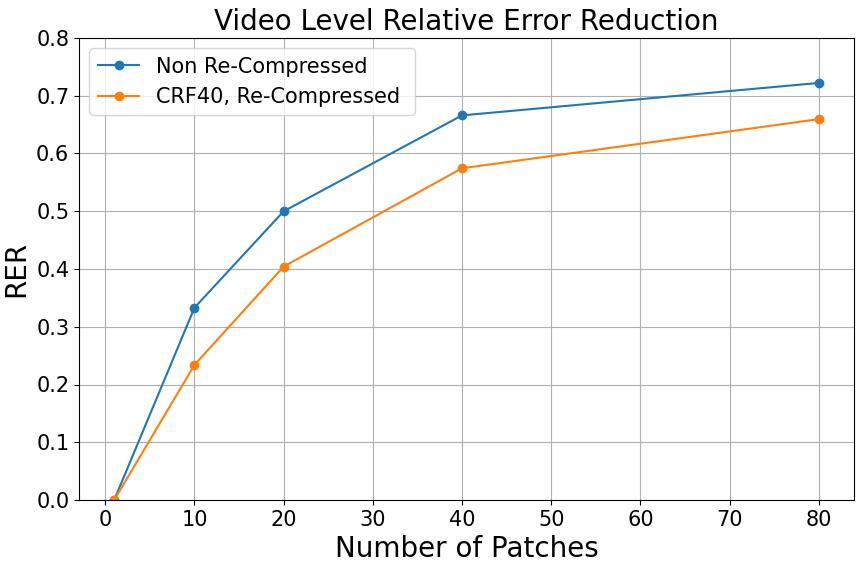}
    \caption{Relative Error Reduction in video-level performance versus frame-level performance of MISLnet over different number of patches used for obtaining video-level detection score.}
    \label{fig:video_rer_plot}
    \pulluppp\pullupp
\end{figure}

The results of these experiments are displayed in Fig.~\ref{fig:video_auc_plot} which shows video-level detection AUC vs number of patches used for detection, and in Fig.~\ref{fig:video_rer_plot} which shows the RER vs. number of patches used for detection.
From these figures, we can see that performance increases as more patches are used to perform video-level detection.
This is particularly strong for re-compressed video, where the AUC grows from 0.953 to 0.984 as 80 patches are used, corresponding to a RER of 66\% over the patch-level detector.
%
These results show that video-level detection can achieve important performance gains by leveraging traces throughout an entire video.

We also note that while the increase in AUC for non-recompressed videos is not large in absolute terms, the RER achieved is substantial.  Specifically, when 80 patches are used, the video-level detector achieves an RER of $72.2\%$ over the patch-level detector.
This is particularly important when performing synthetic video detection at scale, such as if a social media company were to examine all videos uploaded to its service.  Because of the large  number of videos examined, small gains in performance can correspond to a large reduction in the  number of false alarms.


\section{Detection Transferability to New Generators}

New synthetic video generator architectures and generation approaches are emerging at a rapidly. Hence, it is important to understand the transferability of synthetic video detectors to new generators as we know this could be achieved as demonstrated by\cite{mayer2018learning}. 
In this section, we conduct a series of experiments to examine detector transferability in both zero-shot and few-shot transfer scenarios. 


\definecolor{Mercury}{rgb}{0.913,0.913,0.913}
\begin{table}[!h]
    \centering
    \resizebox{0.77\linewidth}{!}{%
    \begin{tblr}{
        width = \linewidth,
        colspec = {m{32mm}m{19mm}m{19mm}},
        row{2} = {c},
        row{6} = {Mercury},
        cell{1}{1} = {r=2}{},
        cell{1}{2} = {c=2}{0.402\linewidth,c},
        cell{3}{2} = {c},
        cell{3}{3} = {c},
        cell{4}{2} = {c},
        cell{4}{3} = {c},
        cell{5}{2} = {c},
        cell{5}{3} = {c},
        cell{6}{2} = {c},
        cell{6}{3} = {c},
        vline{1,2,4} = {-}{},
        hline{1,3,6-7} = {-}{},
        hline{2} = {2-3}{},
    }
        {\textbf{Out-of-distribution}\\\textbf{Generation Method}} & \textbf{Test-only} & \\
        & \textbf{\# Videos} & \textbf{\# Frames}\\
        Sora~\cite{sora} & 38 & 5,000\\
        Pika~\cite{Pika} & 163 & 10,000\\
        VC-v2~\cite{videocrafter2} & 200 & 10,000\\
        \textbf{Total} & 401 & 25,000
    \end{tblr}
    }
    \caption{Statistics of the out-of-distribtion, test-only synthetic video dataset used in Sec.~\ref{sec:zero_few_shot_transfer}.}
    \label{tab:ood_dataset_stats}
    \pulluppp
\end{table}
\label{sec:zero_few_shot_transfer}


\subsection{Zero-Shot Transferability}
\label{subsec:exp_zero_shot}

In our first set of experiment, we examined synthetic video detectors' zero-shot transferability performance.  This corresponds to a detector's ability to detect videos from a new generator without any re-training. 

\begin{table}[!t]
	\centering
	\resizebox{0.85\linewidth}{!}{%
		\begin{tblr}{
				width = \linewidth,
				colspec = {m{35mm}m{20mm}m{25mm}},
				column{2} = {c},
				column{3} = {c},
				vlines,
				hline{1-2,9,6} = {-}{},
			}
			\textbf{Generation Method} & \textbf{Seen Sources} & \textbf{Unseen Source}\\
			VideoCrafter v1\cite{videocrafter1} & 0.993 & 0.773\\
			Cogvideo\cite{cogvideo} & 0.990 & 0.671\\
			Luma\cite{Luma} & 0.991 & 0.702\\
			SVD\cite{stablevideo} & 0.985 & 0.760\\
		\end{tblr}
	}
    \caption{Zero-shot detection performance of MISLnet~\cite{openset}, which was trained on 3 out of 4 synthetic video generation sources and test on the remaining one. Performance numbers are in AUC.}
    \label{tab:exp_leave_one_out}
    \pulluppp\pullupp
\end{table}

We began by training the best performing detector (MISLnet) using data from three of the generators in our training set. We benchmarked the detector's performance on the portion of the test dataset corresponding to generators seen during training. Then, we measured the detector's zero-shot transferability by using it to detect  videos generated by a generator not included in training, i.e. unseen sources.

Results from this experiment are presented in Table~\ref{tab:exp_leave_one_out}.  
These results show that while the detector achieves strong performance on videos from generators seen during training, performance drops significantly when evaluating on new generators.  Specifically, the AUC drops from an average of 0.990 for ``seen'' generators to an average of 0.727 for new, ``unseen'' generators.  

We conducted a similar experiment, in which we used the detector trained on all four generators in our training set to detect synthetic videos from the generators in our Out-of-Distribution testing set described in Section~\ref{subsec:common_video_exp_setup}. These videos were generated using three different generators: Sora, Pika, and VideoCrafter v2.

The results of this experiment are presented in the second colum of Table~\ref{tab:few_shot_learning}, labeled ``Zero-Shot''.  These experiments yielded similar results, in which the detector had significant difficulty detecting videos from these new, unseen generators. 
The only notable exception to this was VideoCrafter v2, with a generator closely similar to VideoCrafter v1.

The results of these experiments are somewhat unsurprising.  As we can see from the generator trace visualizations in Fig.~\ref{fig:trace_visualization}, traces left by different generators can vary substantially.  Furthermore, we know that both video generator architectures and generation approaches (i.e. NeRF, diffusion, transformer, etc.) vary significantly from generator to generator.  As a result, it is  difficult for a detector to capture traces from new generation that leave different traces than those seen in training. 
Plus, this aligns with similar findings obtained for synthetic image detection~\cite{luisa}. 




\subsection{Transferabilty Through Few-Shot Learning}
\label{subsec:exp_few_shot}


Next, we examined the ability of a synthetic video detector to detect a new generator through few-shot learning. 



We began the experiments with a pre-trained detector that identified all four generators in the training set, utilizing MISLnet. We then fine-tuned the detector to recognize each generator in the Out-of-Distribution testing set, using less than one minute of video from each generator. Notably, the fine-tuning videos were not part of the testing set. Subsequently, we evaluated the updated detector's performance on each new generator in the Out-of-Distribution testing set.

Results of this experiment are shown in the second column of Table~\ref{tab:few_shot_learning} from the right, titled Few-Shot.  These results show that the detector can very accurately transfer to detect new generators through few-shot learning.  In each case, the detector achieves an AUC of 0.98 or higher.  This is a substantial increase over the AUCs achieved by zero-shot transferability.  

\begin{table}[!t]
	\centering
	\resizebox{0.86\linewidth}{!}{%
		\begin{tblr}{
				width = \linewidth,
				colspec = {m{30mm}m{16mm}m{15mm}m{13mm}},
				column{2,3,4} = {c},
				vlines,
				hline{1-2,5} = {-}{},
			}
			\textbf{Generation Method} & \textbf{Zero-Shot} & \textbf{Few-Shot} & \textbf{RER}\\
            Sora~\cite{sora} & 0.530 & 0.982 & 96.2\%\\
			Pika~\cite{Pika} & 0.620 & 0.989 & 97.1\%\\
			VideoCrafter v2~\cite{videocrafter2} & 0.939 & 0.996 & 93.4\%
		\end{tblr}
	}
    \caption{Zero-Shot and Few-Shot detection performance of MISLnet~\cite{openset}, which was trained on all training generators, and tested on new generation sources. Performance numbers are measured using AUC and RER.}
    \label{tab:few_shot_learning}
    \pulluppp\pullupp
\end{table}

Notably, for Sora we increase the AUC from 0.530 to 0.982 through few-shot learning.  This corresponds to an AUC boost of 0.452 and an RER of 96\%. 
These results show that synthetic video detectors can be transferred to reliably detect new generators through few-shot learning.  This is particularly important given the rapid pace with which new generators such as Sora are emerging.

\section{Conclusion}

Our paper highlights the challenges of detecting synthetically generated videos, showing that forensic traces in synthetic images and videos differ significantly. Leading to the poor performance of existing synthetic image detectors on AI-generated videos. However, we also showed that it is possible to learn synthetic video traces through the process of training. Resulting in strong and robust detection and attribution using existing synthetic image detectors' architectures. Additionally, we showed that while these detectors have difficulties directly transferring to an unseen generator, strong performance is attainable using very little data.

\subheader{Acknowledgments} This material is based on research sponsored by DARPA and the Air Force Research Laboratory (AFRL) under agreement number HR0011-20-C-0126 and by the National Science Foundation under Award No. 2320600.

{
    \small
    \bibliographystyle{ieeenat_fullname}
    \bibliography{main}

\begin{thebibliography}{98}
\providecommand{\natexlab}[1]{#1}
\providecommand{\url}[1]{\texttt{#1}}
\expandafter\ifx\csname urlstyle\endcsname\relax
  \providecommand{\doi}[1]{doi: #1}\else
  \providecommand{\doi}{doi: \begingroup \urlstyle{rm}\Url}\fi

\bibitem[AI()]{Luma}
Luma AI.
\newblock https://lumalabs.ai/.

\bibitem[Bammey(2024)]{diff1}
Quentin Bammey.
\newblock Synthbuster: Towards detection of diffusion model generated images.
\newblock \emph{IEEE Open Journal of Signal Processing}, 5:\penalty0 1--9,
  2024.

\bibitem[Baraheem and Nguyen(2023)]{xception_method3}
Samah~S Baraheem and Tam~V Nguyen.
\newblock Ai vs. ai: Can ai detect ai-generated images?
\newblock \emph{Journal of Imaging}, 9\penalty0 (10):\penalty0 199, 2023.

\bibitem[Batzolis et~al.(2021)Batzolis, Stanczuk, Schönlieb, and
  Etmann]{controlnet}
Georgios Batzolis, Jan Stanczuk, Carola-Bibiane Schönlieb, and Christian
  Etmann.
\newblock Conditional image generation with score-based diffusion models, 2021.

\bibitem[Bayar and Stamm(2016)]{bayar2016deep}
Belhassen Bayar and Matthew~C Stamm.
\newblock A deep learning approach to universal image manipulation detection
  using a new convolutional layer.
\newblock In \emph{Proceedings of the 4th ACM workshop on information hiding
  and multimedia security}, pages 5--10, 2016.

\bibitem[Bayar and Stamm(2017)]{compression1}
Belhassen Bayar and Matthew~C Stamm.
\newblock Augmented convolutional feature maps for robust cnn-based camera
  model identification.
\newblock In \emph{2017 IEEE International Conference on Image Processing
  (ICIP)}, pages 4098--4102. IEEE, 2017.

\bibitem[Bayar and Stamm(2018)]{bayar2018constrained}
Belhassen Bayar and Matthew~C Stamm.
\newblock Constrained convolutional neural networks: A new approach towards
  general purpose image manipulation detection.
\newblock \emph{IEEE Transactions on Information Forensics and Security},
  13\penalty0 (11):\penalty0 2691--2706, 2018.

\bibitem[Blattmann et~al.(2023)Blattmann, Dockhorn, Kulal, Mendelevitch,
  Kilian, Lorenz, Levi, English, Voleti, Letts, et~al.]{stablevideo}
Andreas Blattmann, Tim Dockhorn, Sumith Kulal, Daniel Mendelevitch, Maciej
  Kilian, Dominik Lorenz, Yam Levi, Zion English, Vikram Voleti, Adam Letts,
  et~al.
\newblock Stable video diffusion: Scaling latent video diffusion models to
  large datasets.
\newblock \emph{arXiv preprint arXiv:2311.15127}, 2023.

\bibitem[Brooks et~al.(2024)Brooks, Peebles, Holmes, DePue, Guo, Jing, Schnurr,
  Taylor, Luhman, Luhman, Ng, Wang, and Ramesh]{sora}
Tim Brooks, Bill Peebles, Connor Holmes, Will DePue, Yufei Guo, Li Jing, David
  Schnurr, Joe Taylor, Troy Luhman, Eric Luhman, Clarence Ng, Ricky Wang, and
  Aditya Ramesh.
\newblock Video generation models as world simulators.
\newblock 2024.

\bibitem[Carvalho et~al.(2017)Carvalho, de~Rezende, Alves, Balieiro, and
  Sovat]{gan5}
Tiago Carvalho, Edmar R.~S. de Rezende, Matheus T.~P. Alves, Fernanda K.~C.
  Balieiro, and Ricardo~B. Sovat.
\newblock Exposing computer generated images by eye’s region classification
  via transfer learning of vgg19 cnn.
\newblock In \emph{2017 16th IEEE International Conference on Machine Learning
  and Applications (ICMLA)}, pages 866--870, 2017.

\bibitem[Chang et~al.(2023)Chang, Zhang, Barber, Maschinot, Lezama, Jiang,
  Yang, Murphy, Freeman, Rubinstein, et~al.]{transformer6}
Huiwen Chang, Han Zhang, Jarred Barber, AJ Maschinot, Jose Lezama, Lu Jiang,
  Ming-Hsuan Yang, Kevin Murphy, William~T Freeman, Michael Rubinstein, et~al.
\newblock Muse: Text-to-image generation via masked generative transformers.
\newblock \emph{arXiv preprint arXiv:2301.00704}, 2023.

\bibitem[Chen et~al.(2021)Chen, Ju, Xiao, Ding, Zheng, and
  de~Albuquerque]{xception_method}
Beijing Chen, Xingwang Ju, Bin Xiao, Weiping Ding, Yuhui Zheng, and Victor
  Hugo~C de Albuquerque.
\newblock Locally gan-generated face detection based on an improved xception.
\newblock \emph{Information Sciences}, 572:\penalty0 16--28, 2021.

\bibitem[Chen et~al.(2023)Chen, Xia, He, Zhang, Cun, Yang, Xing, Liu, Chen,
  Wang, Weng, and Shan]{videocrafter1}
Haoxin Chen, Menghan Xia, Yingqing He, Yong Zhang, Xiaodong Cun, Shaoshu Yang,
  Jinbo Xing, Yaofang Liu, Qifeng Chen, Xintao Wang, Chao Weng, and Ying Shan.
\newblock Videocrafter1: Open diffusion models for high-quality video
  generation, 2023.

\bibitem[Chen et~al.(2024{\natexlab{a}})Chen, Zhang, Cun, Xia, Wang, Weng, and
  Shan]{videocrafter2}
Haoxin Chen, Yong Zhang, Xiaodong Cun, Menghan Xia, Xintao Wang, Chao Weng, and
  Ying Shan.
\newblock Videocrafter2: Overcoming data limitations for high-quality video
  diffusion models, 2024{\natexlab{a}}.

\bibitem[Chen et~al.(2024{\natexlab{b}})Chen, Yao, and Niu]{swin_method2}
Jiaxuan Chen, Jieteng Yao, and Li Niu.
\newblock A single simple patch is all you need for ai-generated image
  detection.
\newblock \emph{arXiv preprint arXiv:2402.01123}, 2024{\natexlab{b}}.

\bibitem[Cheng et~al.(2024)Cheng, Guo, Wang, Nie, and
  Kankanhalli]{xception_method4}
Harry Cheng, Yangyang Guo, Tianyi Wang, Liqiang Nie, and Mohan Kankanhalli.
\newblock Diffusion facial forgery detection.
\newblock \emph{arXiv preprint arXiv:2401.15859}, 2024.

\bibitem[Choi et~al.(2018)Choi, Choi, Kim, Ha, Kim, and Choo]{stargan}
Yunjey Choi, Minje Choi, Munyoung Kim, Jung-Woo Ha, Sunghun Kim, and Jaegul
  Choo.
\newblock Stargan: Unified generative adversarial networks for multi-domain
  image-to-image translation.
\newblock In \emph{Proceedings of the IEEE conference on computer vision and
  pattern recognition}, pages 8789--8797, 2018.

\bibitem[Chollet(2017)]{xception_archit}
Fran{\c{c}}ois Chollet.
\newblock Xception: Deep learning with depthwise separable convolutions.
\newblock In \emph{Proceedings of the IEEE conference on computer vision and
  pattern recognition}, pages 1251--1258, 2017.

\bibitem[Corvi et~al.(2023)Corvi, Cozzolino, Zingarini, Poggi, Nagano, and
  Verdoliva]{luisa}
Riccardo Corvi, Davide Cozzolino, Giada Zingarini, Giovanni Poggi, Koki Nagano,
  and Luisa Verdoliva.
\newblock On the detection of synthetic images generated by diffusion models.
\newblock In \emph{ICASSP 2023-2023 IEEE International Conference on Acoustics,
  Speech and Signal Processing (ICASSP)}, pages 1--5. IEEE, 2023.

\bibitem[Dhariwal and Nichol(2021)]{diff_beat_gans}
Prafulla Dhariwal and Alexander Nichol.
\newblock Diffusion models beat gans on image synthesis.
\newblock In \emph{Advances in Neural Information Processing Systems}, pages
  8780--8794. Curran Associates, Inc., 2021.

\bibitem[Esser et~al.(2021)Esser, Rombach, and Ommer]{transformer3}
Patrick Esser, Robin Rombach, and Bjorn Ommer.
\newblock Taming transformers for high-resolution image synthesis.
\newblock In \emph{Proceedings of the IEEE/CVF conference on computer vision
  and pattern recognition}, pages 12873--12883, 2021.

\bibitem[Fahn and Wu(2022)]{resnet34_method2}
Chin-Shyurng Fahn and Tzu-Chin Wu.
\newblock A deep-neural-network-based approach to detecting forgery images
  generated from various generative adversarial networks.
\newblock In \emph{2022 International Conference on Machine Learning and
  Cybernetics (ICMLC)}, pages 115--123, 2022.

\bibitem[Fang et~al.(2023)Fang, Nguyen, and Stamm]{openset}
Shengbang Fang, Tai~D Nguyen, and Matthew~C Stamm.
\newblock Open set synthetic image source attribution.
\newblock In \emph{34th British Machine Vision Conference 2023, {BMVC} 2023,
  Aberdeen, UK, November 20-24, 2023}. BMVA, 2023.

\bibitem[Fotedar and Wang(2019)]{transformer1}
NA Fotedar and JH Wang.
\newblock Bumblebee: Text-to-image generation with transformers.
\newblock In \emph{Proceedings of the 2019 IEEE International Conference on
  Image Processing (ICIP)}, pages 3465--3469, 2019.

\bibitem[Ge et~al.(2022)Ge, Hayes, Yang, Yin, Pang, Jacobs, Huang, and
  Parikh]{trans1}
Songwei Ge, Thomas Hayes, Harry Yang, Xi Yin, Guan Pang, David Jacobs, Jia-Bin
  Huang, and Devi Parikh.
\newblock Long video generation with time-agnostic vqgan and time-sensitive
  transformer.
\newblock In \emph{European Conference on Computer Vision}, pages 102--118.
  Springer, 2022.

\bibitem[Goebel et~al.(2020)Goebel, Nataraj, Nanjundaswamy, Mohammed,
  Chandrasekaran, and Manjunath]{vgg16_method}
Michael Goebel, Lakshmanan Nataraj, Tejaswi Nanjundaswamy, Tajuddin~Manhar
  Mohammed, Shivkumar Chandrasekaran, and BS Manjunath.
\newblock Detection, attribution and localization of gan generated images.
\newblock \emph{arXiv preprint arXiv:2007.10466}, 2020.

\bibitem[Goodfellow et~al.(2014)Goodfellow, Pouget-Abadie, Mirza, Xu,
  Warde-Farley, Ozair, Courville, and Bengio]{goodfellow2014generative}
Ian Goodfellow, Jean Pouget-Abadie, Mehdi Mirza, Bing Xu, David Warde-Farley,
  Sherjil Ozair, Aaron Courville, and Yoshua Bengio.
\newblock Generative adversarial nets.
\newblock \emph{Advances in neural information processing systems}, 27, 2014.

\bibitem[Guo et~al.(2021)Guo, Yang, Chen, and Sun]{mislnet_method1}
Zhiqing Guo, Gaobo Yang, Jiyou Chen, and Xingming Sun.
\newblock Fake face detection via adaptive manipulation traces extraction
  network.
\newblock \emph{Computer Vision and Image Understanding}, 204:\penalty0 103170,
  2021.

\bibitem[Haji-Ali et~al.(2023)Haji-Ali, Balakrishnan, and
  Ordonez]{haji2023elasticdiffusion}
Moayed Haji-Ali, Guha Balakrishnan, and Vicente Ordonez.
\newblock Elasticdiffusion: Training-free arbitrary size image generation.
\newblock \emph{arXiv preprint arXiv:2311.18822}, 2023.

\bibitem[Hatamizadeh et~al.(2023)Hatamizadeh, Song, Liu, Kautz, and
  Vahdat]{transformer5}
Ali Hatamizadeh, Jiaming Song, Guilin Liu, Jan Kautz, and Arash Vahdat.
\newblock Diffit: Diffusion vision transformers for image generation.
\newblock \emph{arXiv preprint arXiv:2312.02139}, 2023.

\bibitem[He et~al.(2016)He, Zhang, Ren, and Sun]{resnet_archit}
Kaiming He, Xiangyu Zhang, Shaoqing Ren, and Jian Sun.
\newblock Deep residual learning for image recognition.
\newblock In \emph{Proceedings of the IEEE conference on computer vision and
  pattern recognition}, pages 770--778, 2016.

\bibitem[Ho et~al.(2020)Ho, Jain, and Abbeel]{ho2020denoising}
Jonathan Ho, Ajay Jain, and Pieter Abbeel.
\newblock Denoising diffusion probabilistic models.
\newblock \emph{Advances in Neural Information Processing Systems},
  33:\penalty0 6840--6851, 2020.

\bibitem[Ho et~al.(2022)Ho, Saharia, Chan, Fleet, Norouzi, and
  Salimans]{ho2022cascaded}
Jonathan Ho, Chitwan Saharia, William Chan, David~J Fleet, Mohammad Norouzi,
  and Tim Salimans.
\newblock Cascaded diffusion models for high fidelity image generation.
\newblock \emph{J. Mach. Learn. Res.}, 23\penalty0 (47):\penalty0 1--33, 2022.

\bibitem[Hong et~al.(2022)Hong, Ding, Zheng, Liu, and Tang]{cogvideo}
Wenyi Hong, Ming Ding, Wendi Zheng, Xinghan Liu, and Jie Tang.
\newblock Cogvideo: Large-scale pretraining for text-to-video generation via
  transformers.
\newblock \emph{arXiv preprint arXiv:2205.15868}, 2022.

\bibitem[Hosler et~al.(2019{\natexlab{a}})Hosler, Mayer, Bayar, Zhao, Chen,
  Shackleford, and Stamm]{hosler2019video}
Brian Hosler, Owen Mayer, Belhassen Bayar, Xinwei Zhao, Chen Chen, James~A
  Shackleford, and Matthew~Christopher Stamm.
\newblock A video camera model identification system using deep learning and
  fusion.
\newblock In \emph{ICASSP 2019-2019 IEEE international conference on acoustics,
  speech and signal processing (ICASSP)}, pages 8271--8275. IEEE,
  2019{\natexlab{a}}.

\bibitem[Hosler et~al.(2019{\natexlab{b}})Hosler, Zhao, Mayer, Chen,
  Shackleford, and Stamm]{videoacid}
Brian~C Hosler, Xinwei Zhao, Owen Mayer, Chen Chen, James~A Shackleford, and
  Matthew~C Stamm.
\newblock The video authentication and camera identification database: A new
  database for video forensics.
\newblock \emph{IEEE Access}, 7:\penalty0 76937--76948, 2019{\natexlab{b}}.

\bibitem[Huang et~al.(2017)Huang, Liu, Van Der~Maaten, and
  Weinberger]{densenet_archit}
Gao Huang, Zhuang Liu, Laurens Van Der~Maaten, and Kilian~Q Weinberger.
\newblock Densely connected convolutional networks.
\newblock In \emph{Proceedings of the IEEE conference on computer vision and
  pattern recognition}, pages 4700--4708, 2017.

\bibitem[Hulzebosch et~al.(2020)Hulzebosch, Ibrahimi, and
  Worring]{xception_method2}
Nils Hulzebosch, Sarah Ibrahimi, and Marcel Worring.
\newblock Detecting cnn-generated facial images in real-world scenarios.
\newblock In \emph{Proceedings of the IEEE/CVF Conference on Computer Vision
  and Pattern Recognition (CVPR) Workshops}, 2020.

\bibitem[Jiang et~al.(2023)Jiang, Yang, Koh, Wu, Loy, and Liu]{trans3}
Yuming Jiang, Shuai Yang, Tong~Liang Koh, Wayne Wu, Chen~Change Loy, and Ziwei
  Liu.
\newblock Text2performer: Text-driven human video generation.
\newblock In \emph{Proceedings of the IEEE/CVF International Conference on
  Computer Vision (ICCV)}, pages 22747--22757, 2023.

\bibitem[Karras et~al.(2018)Karras, Aila, Laine, and Lehtinen]{progan}
Tero Karras, Timo Aila, Samuli Laine, and Jaakko Lehtinen.
\newblock Progressive growing of gans for improved quality, stability, and
  variation.
\newblock In \emph{International Conference on Learning Representations}, 2018.

\bibitem[Karras et~al.(2019)Karras, Laine, and Aila]{stylegan}
Tero Karras, Samuli Laine, and Timo Aila.
\newblock A style-based generator architecture for generative adversarial
  networks.
\newblock In \emph{Proceedings of the IEEE/CVF conference on computer vision
  and pattern recognition}, pages 4401--4410, 2019.

\bibitem[Karras et~al.(2020)Karras, Laine, Aittala, Hellsten, Lehtinen, and
  Aila]{stylegan2}
Tero Karras, Samuli Laine, Miika Aittala, Janne Hellsten, Jaakko Lehtinen, and
  Timo Aila.
\newblock Analyzing and improving the image quality of stylegan.
\newblock In \emph{Proceedings of the IEEE/CVF conference on computer vision
  and pattern recognition}, pages 8110--8119, 2020.

\bibitem[Karras et~al.(2021)Karras, Aittala, Laine, H{\"a}rk{\"o}~nen,
  Hellsten, Lehtinen, and Aila]{stylegan3}
Tero Karras, Miika Aittala, Samuli Laine, Erik H{\"a}rk{\"o}~nen, Janne
  Hellsten, Jaakko Lehtinen, and Timo Aila.
\newblock Alias-free generative adversarial networks.
\newblock \emph{Advances in Neural Information Processing Systems},
  34:\penalty0 852--863, 2021.

\bibitem[Kirchner and Fridrich(2010)]{compression2}
Matthias Kirchner and Jessica Fridrich.
\newblock On detection of median filtering in digital images.
\newblock In \emph{Media forensics and security II}, pages 371--382. SPIE,
  2010.

\bibitem[Labs()]{Pika}
Pika Labs.
\newblock https://pika.art/.

\bibitem[Lee et~al.(2021)Lee, Tariq, Shin, and Woo]{densenet_method3}
Sangyup Lee, Shahroz Tariq, Youjin Shin, and Simon~S Woo.
\newblock Detecting handcrafted facial image manipulations and gan-generated
  facial images using shallow-fakefacenet.
\newblock \emph{Applied soft computing}, 105:\penalty0 107256, 2021.

\bibitem[Li et~al.(2018)Li, Chen, Li, and Tan]{vgg16_method3}
Haodong Li, Han Chen, Bin Li, and Shunquan Tan.
\newblock Can forensic detectors identify gan generated images?
\newblock In \emph{2018 Asia-Pacific Signal and Information Processing
  Association Annual Summit and Conference (APSIPA ASC)}, pages 722--727. IEEE,
  2018.

\bibitem[Li et~al.(2022)Li, He, Li, Wang, and Zhang]{resnet50_method_3}
Weichuang Li, Peisong He, Haoliang Li, Hongxia Wang, and Ruimei Zhang.
\newblock Detection of gan-generated images by estimating artifact similarity.
\newblock \emph{IEEE Signal Processing Letters}, 29:\penalty0 862--866, 2022.

\bibitem[Lin et~al.(2014)Lin, Maire, Belongie, Hays, Perona, Ramanan,
  Doll{\'a}r, and Zitnick]{coco}
Tsung-Yi Lin, Michael Maire, Serge Belongie, James Hays, Pietro Perona, Deva
  Ramanan, Piotr Doll{\'a}r, and C~Lawrence Zitnick.
\newblock Microsoft coco: Common objects in context.
\newblock In \emph{Computer Vision--ECCV 2014: 13th European Conference,
  Zurich, Switzerland, September 6-12, 2014, Proceedings, Part V 13}, pages
  740--755. Springer, 2014.

\bibitem[Liu et~al.(2021)Liu, Lin, Cao, Hu, Wei, Zhang, Lin, and Guo]{swint}
Ze Liu, Yutong Lin, Yue Cao, Han Hu, Yixuan Wei, Zheng Zhang, Stephen Lin, and
  Baining Guo.
\newblock Swin transformer: Hierarchical vision transformer using shifted
  windows.
\newblock In \emph{Proceedings of the IEEE/CVF international conference on
  computer vision}, pages 10012--10022, 2021.

\bibitem[Lorenz et~al.(2023)Lorenz, Durall, and Keuper]{diff2}
Peter Lorenz, Ricard~L. Durall, and Janis Keuper.
\newblock Detecting images generated by deep diffusion models using their local
  intrinsic dimensionality.
\newblock In \emph{Proceedings of the IEEE/CVF International Conference on
  Computer Vision (ICCV) Workshops}, pages 448--459, 2023.

\bibitem[Ma et~al.(2024)Ma, Wang, Jia, Chen, Liu, Li, Chen, and Qiao]{trans7}
Xin Ma, Yaohui Wang, Gengyun Jia, Xinyuan Chen, Ziwei Liu, Yuan-Fang Li,
  Cunjian Chen, and Yu Qiao.
\newblock Latte: Latent diffusion transformer for video generation.
\newblock \emph{arXiv preprint arXiv:2401.03048}, 2024.

\bibitem[Malhotra(2021)]{resnet34_method3}
Yishu Malhotra.
\newblock Image forgery detection using textural features and deep learning.
\newblock 2021.

\bibitem[Marra et~al.(2018)Marra, Gragnaniello, Cozzolino, and
  Verdoliva]{densenet_method}
Francesco Marra, Diego Gragnaniello, Davide Cozzolino, and Luisa Verdoliva.
\newblock Detection of gan-generated fake images over social networks.
\newblock In \emph{2018 IEEE conference on multimedia information processing
  and retrieval (MIPR)}, pages 384--389. IEEE, 2018.

\bibitem[Marra et~al.(2019)Marra, Gragnaniello, Verdoliva, and Poggi]{gan1}
Francesco Marra, Diego Gragnaniello, Luisa Verdoliva, and Giovanni Poggi.
\newblock Do gans leave artificial fingerprints?
\newblock In \emph{2019 IEEE Conference on Multimedia Information Processing
  and Retrieval (MIPR)}, pages 506--511, 2019.

\bibitem[May et~al.(2023)May, Trapeznikov, Fang, and
  Stamm]{may2023comprehensive}
Brandon~B May, Kirill Trapeznikov, Shengbang Fang, and Matthew Stamm.
\newblock Comprehensive dataset of synthetic and manipulated overhead imagery
  for development and evaluation of forensic tools.
\newblock In \emph{Proceedings of the 2023 ACM Workshop on Information Hiding
  and Multimedia Security}, pages 145--150, 2023.

\bibitem[Mayer et~al.(2018)Mayer, Bayar, and Stamm]{mayer2018learning}
Owen Mayer, Belhassen Bayar, and Matthew~C Stamm.
\newblock Learning unified deep-features for multiple forensic tasks.
\newblock In \emph{Proceedings of the 6th ACM workshop on information hiding
  and multimedia security}, pages 79--84, 2018.

\bibitem[Mayer et~al.(2020)Mayer, Hosler, and Stamm]{mayer2020open}
Owen Mayer, Brian Hosler, and Matthew~C Stamm.
\newblock Open set video camera model verification.
\newblock In \emph{ICASSP 2020-2020 IEEE international conference on acoustics,
  speech and signal processing (ICASSP)}, pages 2962--2966. IEEE, 2020.

\bibitem[Meena and Tyagi(2020)]{densenet_method2}
Kunj~Bihari Meena and Vipin Tyagi.
\newblock A deep learning based method to discriminate between photorealistic
  computer generated images and photographic images.
\newblock In \emph{Advances in Computing and Data Sciences: 4th International
  Conference, ICACDS 2020, Valletta, Malta, April 24--25, 2020, Revised
  Selected Papers 4}, pages 212--223. Springer, 2020.

\bibitem[Mi et~al.(2020)Mi, Jiang, Sun, and Xu]{gan9}
Zhongjie Mi, Xinghao Jiang, Tanfeng Sun, and Ke Xu.
\newblock Gan-generated image detection with self-attention mechanism against
  gan generator defect.
\newblock \emph{IEEE Journal of Selected Topics in Signal Processing},
  14\penalty0 (5):\penalty0 969--981, 2020.

\bibitem[Midjourney()]{Midjourney}
Midjourney.
\newblock Midjourney ai - free image generator.

\bibitem[ML()]{Runway}
Runway ML.
\newblock Advancing creativity with artificial intelligence.

\bibitem[Monfort et~al.(2021)Monfort, Jin, Liu, Harwath, Feris, Glass, and
  Oliva]{mit}
Mathew Monfort, SouYoung Jin, Alexander Liu, David Harwath, Rogerio Feris,
  James Glass, and Aude Oliva.
\newblock Spoken moments: Learning joint audio-visual representations from
  video descriptions.
\newblock In \emph{Proceedings of the IEEE/CVF Conference on Computer Vision
  and Pattern Recognition (CVPR)}, pages 14871--14881, 2021.

\bibitem[Monjezi et~al.(2023)Monjezi, Trivedi, Tan, and Tizpaz-Niari]{monjezi}
Verya Monjezi, Ashutosh Trivedi, Gang Tan, and Saeid Tizpaz-Niari.
\newblock Information-theoretic testing and debugging of fairness defects in
  deep neural networks.
\newblock In \emph{2023 IEEE/ACM 45th International Conference on Software
  Engineering (ICSE)}, pages 1571--1582. IEEE, 2023.

\bibitem[Nataraj et~al.(2019)Nataraj, Mohammed, Chandrasekaran, Flenner, Bappy,
  Roy-Chowdhury, and Manjunath]{gan7}
Lakshmanan Nataraj, Tajuddin~Manhar Mohammed, Shivkumar Chandrasekaran, Arjuna
  Flenner, Jawadul~H Bappy, Amit~K Roy-Chowdhury, and BS Manjunath.
\newblock Detecting gan generated fake images using co-occurrence matrices.
\newblock \emph{arXiv preprint arXiv:1903.06836}, 2019.

\bibitem[Nguyen et~al.(2024)Nguyen, Fang, and Stamm]{nguyen2024videofact}
Tai~D Nguyen, Shengbang Fang, and Matthew~C Stamm.
\newblock Videofact: Detecting video forgeries using attention, scene context,
  and forensic traces.
\newblock In \emph{Proceedings of the IEEE/CVF Winter Conference on
  Applications of Computer Vision}, pages 8563--8573, 2024.

\bibitem[Parmar et~al.(2018)Parmar, Vaswani, Uszkoreit, Kaiser, Shazeer, Ku,
  and Tran]{transformer2}
Niki Parmar, Ashish Vaswani, Jakob Uszkoreit, Lukasz Kaiser, Noam Shazeer,
  Alexander Ku, and Dustin Tran.
\newblock Image transformer.
\newblock In \emph{International conference on machine learning}, pages
  4055--4064. PMLR, 2018.

\bibitem[Pham et~al.(2020)Pham, Dang, Tang, and Nguyen]{vgg16_method2}
Kha-Luan Pham, Khanh-Mai Dang, Loi-Phat Tang, and Thanh-Nhan Nguyen.
\newblock Gan generated portraits detection using modified vgg-16 and
  efficientnet.
\newblock In \emph{2020 7th NAFOSTED Conference on Information and Computer
  Science (NICS)}, pages 344--349. IEEE, 2020.

\bibitem[Quan et~al.(2018)Quan, Wang, Yan, and Zhang]{gan4}
Weize Quan, Kai Wang, Dong-Ming Yan, and Xiaopeng Zhang.
\newblock Distinguishing between natural and computer-generated images using
  convolutional neural networks.
\newblock \emph{IEEE Transactions on Information Forensics and Security},
  13\penalty0 (11):\penalty0 2772--2787, 2018.

\bibitem[Quan et~al.(2024)Quan, Deng, Wang, and Yan]{diff4}
Weize Quan, Pengfei Deng, Kai Wang, and Dong-Ming Yan.
\newblock Cgformer: Vit-based network for identifying computer-generated images
  with token labeling.
\newblock \emph{IEEE Transactions on Information Forensics and Security},
  19:\penalty0 235--250, 2024.

\bibitem[Rakhimov et~al.(2020)Rakhimov, Volkhonskiy, Artemov, Zorin, and
  Burnaev]{trans4}
Ruslan Rakhimov, Denis Volkhonskiy, Alexey Artemov, Denis Zorin, and Evgeny
  Burnaev.
\newblock Latent video transformer.
\newblock \emph{arXiv preprint arXiv:2006.10704}, 2020.

\bibitem[Ramesh et~al.(2022)Ramesh, Dhariwal, Nichol, Chu, and Chen]{dalle2}
Aditya Ramesh, Prafulla Dhariwal, Alex Nichol, Casey Chu, and Mark Chen.
\newblock Hierarchical text-conditional image generation with clip latents.
\newblock \emph{arXiv preprint arXiv:2204.06125}, 1\penalty0 (2):\penalty0 3,
  2022.

\bibitem[Rauniyar et~al.(2023)Rauniyar, Raj, Kumar, Kandu, Singh, and
  Gupta]{diffusion7}
Apoorva Rauniyar, Aryan Raj, Ashish Kumar, Ashish~Kumar Kandu, Astha Singh, and
  Anjani Gupta.
\newblock Text to image generator with latent diffusion models.
\newblock In \emph{2023 International Conference on Computational Intelligence,
  Communication Technology and Networking (CICTN)}, pages 144--148, 2023.

\bibitem[Rombach et~al.(2022)Rombach, Blattmann, Lorenz, Esser, and
  Ommer]{stablediffusion}
Robin Rombach, Andreas Blattmann, Dominik Lorenz, Patrick Esser, and Bj{\"o}rn
  Ommer.
\newblock High-resolution image synthesis with latent diffusion models.
\newblock In \emph{Proceedings of the IEEE/CVF conference on computer vision
  and pattern recognition}, pages 10684--10695, 2022.

\bibitem[Simonyan and Zisserman(2014)]{vgg16_archit}
Karen Simonyan and Andrew Zisserman.
\newblock Very deep convolutional networks for large-scale image recognition.
\newblock \emph{arXiv preprint arXiv:1409.1556}, 2014.

\bibitem[Singer et~al.(2022)Singer, Polyak, Hayes, Yin, An, Zhang, Hu, Yang,
  Ashual, Gafni, et~al.]{trans8}
Uriel Singer, Adam Polyak, Thomas Hayes, Xi Yin, Jie An, Songyang Zhang, Qiyuan
  Hu, Harry Yang, Oron Ashual, Oran Gafni, et~al.
\newblock Make-a-video: Text-to-video generation without text-video data.
\newblock \emph{arXiv preprint arXiv:2209.14792}, 2022.

\bibitem[Sinitsa and Fried(2024)]{dif}
Sergey Sinitsa and Ohad Fried.
\newblock Deep image fingerprint: Towards low budget synthetic image detection
  and model lineage analysis.
\newblock In \emph{Proceedings of the IEEE/CVF Winter Conference on
  Applications of Computer Vision}, pages 4067--4076, 2024.

\bibitem[Snehith.K et~al.(2023)Snehith.K, Anuradha.G, Vemuri, Krishna.N,
  Veerappampalayam~Easwaramoorthy, and Abualigah]{resnet50_method4}
Snehith.K, Anuradha.G, Vemuri, Hari Krishna.N, Sathishkumar
  Veerappampalayam~Easwaramoorthy, and Laith Abualigah.
\newblock \emph{Detection of GAN Generated Fake Satellite Images Using Deep
  Learning}.
\newblock 2023.

\bibitem[Sychandran and Shreelekshmi(2022)]{xception_method5}
CS Sychandran and R Shreelekshmi.
\newblock A hybrid xception-ensemble model for the detection of computer
  generated images.
\newblock In \emph{2022 IEEE International Conference on Artificial
  Intelligence in Engineering and Technology (IICAIET)}, pages 1--6. IEEE,
  2022.

\bibitem[Uccheddu et~al.(2010)Uccheddu, De~Rosa, Piva, and Barni]{compression3}
Francesca Uccheddu, Alessia De~Rosa, Alessandro Piva, and Mauro Barni.
\newblock Detection of resampled images: performance analysis and practical
  challenges.
\newblock In \emph{2010 18th European Signal Processing Conference}, pages
  1675--1679. IEEE, 2010.

\bibitem[Vahdati and Stamm(2023)]{vahdati2023detecting}
Danial~Samadi Vahdati and Matthew~C Stamm.
\newblock Detecting gan-generated synthetic images using semantic
  inconsistencies.
\newblock \emph{Electronic Imaging}, 35:\penalty0 1--6, 2023.

\bibitem[Wang et~al.(2019)Wang, Li, Luo, Shi, and Jha]{gan10}
Jinwei Wang, Ting Li, Xiangyang Luo, Yun-Qing Shi, and Sunil~Kr. Jha.
\newblock Identifying computer generated images based on quaternion central
  moments in color quaternion wavelet domain.
\newblock \emph{IEEE Transactions on Circuits and Systems for Video
  Technology}, 29\penalty0 (9):\penalty0 2775--2785, 2019.

\bibitem[Wang et~al.(2020)Wang, Wang, Zhang, Owens, and Efros]{easydetect}
Sheng-Yu Wang, Oliver Wang, Richard Zhang, Andrew Owens, and Alexei~A Efros.
\newblock Cnn-generated images are surprisingly easy to spot... for now.
\newblock In \emph{Proceedings of the IEEE/CVF conference on computer vision
  and pattern recognition}, pages 8695--8704, 2020.

\bibitem[Xiang et~al.(2023)Xiang, Yang, Huang, and Tong]{im_gen_3d_diff}
Jianfeng Xiang, Jiaolong Yang, Binbin Huang, and Xin Tong.
\newblock 3d-aware image generation using 2d diffusion models.
\newblock In \emph{Proceedings of the IEEE/CVF International Conference on
  Computer Vision (ICCV)}, pages 2383--2393, 2023.

\bibitem[Yadav et~al.(2019)Yadav, Chen, and Ross]{gan11}
Shivangi Yadav, Cunjian Chen, and Arun Ross.
\newblock Synthesizing iris images using rasgan with application in
  presentation attack detection.
\newblock In \emph{Proceedings of the IEEE/CVF Conference on Computer Vision
  and Pattern Recognition (CVPR) Workshops}, 2019.

\bibitem[Yan et~al.(2021)Yan, Zhang, Abbeel, and Srinivas]{trans2}
Wilson Yan, Yunzhi Zhang, Pieter Abbeel, and Aravind Srinivas.
\newblock Videogpt: Video generation using vq-vae and transformers.
\newblock \emph{arXiv preprint arXiv:2104.10157}, 2021.

\bibitem[Yan et~al.(2023)Yan, Hafner, James, and Abbeel]{trans5}
Wilson Yan, Danijar Hafner, Stephen James, and Pieter Abbeel.
\newblock Temporally consistent transformers for video generation.
\newblock In \emph{Proceedings of the 40th International Conference on Machine
  Learning}, pages 39062--39098. PMLR, 2023.

\bibitem[Yang et~al.(2020)Yang, Baracchi, Ni, Zhao, Argenti, and
  Piva]{densenet_method4}
Pengpeng Yang, Daniele Baracchi, Rongrong Ni, Yao Zhao, Fabrizio Argenti, and
  Alessandro Piva.
\newblock A survey of deep learning-based source image forensics.
\newblock \emph{Journal of Imaging}, 6\penalty0 (3):\penalty0 9, 2020.

\bibitem[Yao et~al.(2022)Yao, Zhang, Ni, Shen, Chen, and Xu]{vgg16_method4}
Ye Yao, Zhuxi Zhang, Xuan Ni, Zhangyi Shen, Linqiang Chen, and Dawen Xu.
\newblock Cgnet: Detecting computer-generated images based on transfer learning
  with attention module.
\newblock \emph{Signal Processing: Image Communication}, 105:\penalty0 116692,
  2022.

\bibitem[Yu et~al.(2015)Yu, Seff, Zhang, Song, Funkhouser, and Xiao]{lsun}
Fisher Yu, Ari Seff, Yinda Zhang, Shuran Song, Thomas Funkhouser, and Jianxiong
  Xiao.
\newblock Lsun: Construction of a large-scale image dataset using deep learning
  with humans in the loop.
\newblock \emph{arXiv preprint arXiv:1506.03365}, 2015.

\bibitem[Yu et~al.(2023)Yu, Cheng, Sohn, Lezama, Zhang, Chang, Hauptmann, Yang,
  Hao, Essa, and Jiang]{trans6}
Lijun Yu, Yong Cheng, Kihyuk Sohn, Jos\'e Lezama, Han Zhang, Huiwen Chang,
  Alexander~G. Hauptmann, Ming-Hsuan Yang, Yuan Hao, Irfan Essa, and Lu Jiang.
\newblock Magvit: Masked generative video transformer.
\newblock In \emph{Proceedings of the IEEE/CVF Conference on Computer Vision
  and Pattern Recognition (CVPR)}, pages 10459--10469, 2023.

\bibitem[Zhang et~al.(2022)Zhang, Gu, Zhang, Bao, Chen, Wen, Wang, and
  Guo]{transformer4}
Bowen Zhang, Shuyang Gu, Bo Zhang, Jianmin Bao, Dong Chen, Fang Wen, Yong Wang,
  and Baining Guo.
\newblock Styleswin: Transformer-based gan for high-resolution image
  generation.
\newblock In \emph{Proceedings of the IEEE/CVF conference on computer vision
  and pattern recognition}, pages 11304--11314, 2022.

\bibitem[Zhang et~al.(2019{\natexlab{a}})Zhang, Liang, Zhang, Wang, and
  Li]{gan8}
Kejun Zhang, Yu Liang, Jianyi Zhang, Zhiqiang Wang, and Xinxin Li.
\newblock No one can escape: A general approach to detect tampered and
  generated image.
\newblock \emph{IEEE Access}, 7:\penalty0 129494--129503, 2019{\natexlab{a}}.

\bibitem[Zhang et~al.(2019{\natexlab{b}})Zhang, Karaman, and
  Chang]{resnet34_method}
Xu Zhang, Svebor Karaman, and Shih-Fu Chang.
\newblock Detecting and simulating artifacts in gan fake images.
\newblock In \emph{2019 IEEE international workshop on information forensics
  and security (WIFS)}, pages 1--6. IEEE, 2019{\natexlab{b}}.

\bibitem[Zhao et~al.(2023)Zhao, Wu, Adeke, Qiao, and Wang]{resnet_method_2}
Junjie Zhao, Junfeng Wu, James~Msughter Adeke, Sen Qiao, and Jinwei Wang.
\newblock Detecting high-resolution adversarial images with few-shot deep
  learning.
\newblock \emph{Remote Sensing}, 15\penalty0 (9):\penalty0 2379, 2023.

\bibitem[Zhu et~al.(2017)Zhu, Park, Isola, and Efros]{cyclegan}
Jun-Yan Zhu, Taesung Park, Phillip Isola, and Alexei~A Efros.
\newblock Unpaired image-to-image translation using cycle-consistent
  adversarial networks.
\newblock In \emph{Proceedings of the IEEE international conference on computer
  vision}, pages 2223--2232, 2017.

\bibitem[Zhu et~al.(2023{\natexlab{a}})Zhu, Chen, Huang, Li, Hu, Hu, and
  Wang]{genimage}
Mingjian Zhu, Hanting Chen, Mouxiao Huang, Wei Li, Hailin Hu, Jie Hu, and Yunhe
  Wang.
\newblock Gendet: Towards good generalizations for ai-generated image
  detection.
\newblock \emph{arXiv preprint arXiv:2312.08880}, 2023{\natexlab{a}}.

\bibitem[Zhu et~al.(2023{\natexlab{b}})Zhu, Li, Wang, He, and
  Yao]{cond_text_im_gen_diff}
Yuanzhi Zhu, Zhaohai Li, Tianwei Wang, Mengchao He, and Cong Yao.
\newblock Conditional text image generation with diffusion models.
\newblock In \emph{Proceedings of the IEEE/CVF Conference on Computer Vision
  and Pattern Recognition (CVPR)}, pages 14235--14245, 2023{\natexlab{b}}.

\end{thebibliography}
}


\end{document}